\documentclass{article}
\usepackage[utf8]{inputenc}
\usepackage{microtype}
\usepackage{graphicx}
\usepackage{subfigure}
\usepackage{booktabs}
\usepackage[skip=1ex]{caption}
\usepackage{tabularray}
\UseTblrLibrary{booktabs}
\usepackage{siunitx}
\usepackage[breaklinks=true]{hyperref}
\hypersetup{colorlinks=true,linkcolor=blue, citecolor=blue, urlcolor=blue}
\usepackage{authblk}
\usepackage[backend=biber,
style=numeric,
sorting=ynt]{biblatex}
\usepackage{amsmath}
\usepackage{amssymb}
\usepackage{mathtools}
\usepackage{amsthm}
\usepackage[normalem]{ulem}
\usepackage[capitalize,noabbrev]{cleveref}
\usepackage[textsize=tiny]{todonotes}
\usepackage{cancel}
\usepackage{algorithm}
\usepackage{algorithmic}
\usepackage{xcolor}
\usepackage[british]{babel}
\usepackage{hhline}
\usepackage{multirow,booktabs, makecell}

\theoremstyle{plain}

\theoremstyle{definition}

\theoremstyle{remark}

\newcommand{\ALGCOMRIGHT}[1]{\hfill \textcolor{darkgray}{\#\textrm{#1}}}

\newcolumntype{P}[1]{>{\centering\arraybackslash}p{#1}}
\newcolumntype{M}[1]{>{\centering\arraybackslash}m{#1}}

\DeclareMathOperator{\E}{\mathbb{E}}
\DeclareMathOperator{\V}{\sigma^2}
\DeclareMathOperator{\Q}{\boldsymbol{Q}}
\DeclareMathOperator{\w}{\boldsymbol{w}}
\DeclareMathOperator{\wm}{\boldsymbol{v}}
\DeclareMathOperator{\wf}{\boldsymbol{u}}
\DeclareMathOperator{\m}{\boldsymbol{m}}
\DeclareMathOperator{\M}{\boldsymbol{M}}
\DeclareMathOperator{\F}{\boldsymbol{F}}
\DeclareMathOperator{\Z}{\boldsymbol{Z}}
\DeclareMathOperator{\pred}{\boldsymbol{P}}

\title{Pathspace Kalman Filters with Dynamic Process Uncertainty for Analyzing Time-course Data}

\author[1]{Chaitra Agrahar$^*$}
\author[1]{William Poole\footnote{Represents equal contribution. \\ Correspondence to: cagrahar@altoslabs.com, wpoole@altoslabs.com}}
\author[1]{Simone Bianco}
\author[1]{Hana El-Samad}
\affil{Bay Area Institute of Science, Altos Labs, CA, USA}
\date{\today}

\begin{document}
\maketitle

\begin{abstract}
\noindent Kalman Filter (KF) is an optimal linear state prediction algorithm, with applications in fields as diverse as engineering, economics, robotics, and space exploration. Here, we develop an extension to a non-linear adaptive KF closely related to Bayesian smoothing algorithms, called a Pathspace Kalman Filter (PKF) which allows us to a) dynamically track and update the uncertainties associated with the underlying data and prior knowledge, and b) take as input an entire trajectory and an underlying mechanistic model, and using a Bayesian methodology quantify the different sources of uncertainty. An application of this algorithm is to automatically detect temporal windows where the internal mechanistic model deviates from the data in a time-dependent manner. First, we provide theorems characterizing the convergence of the PKF algorithm. Then, we numerically demonstrate that the PKF outperforms state-of-the-art methods on a synthetic dataset by over an order of magnitude. Finally, we apply this method to biological time-course dataset involving over 1.8 million gene expression measurements.  
\end{abstract}

\section{Introduction}
\begin{figure*}[h!]
    \centering
    \includegraphics[width=\linewidth]{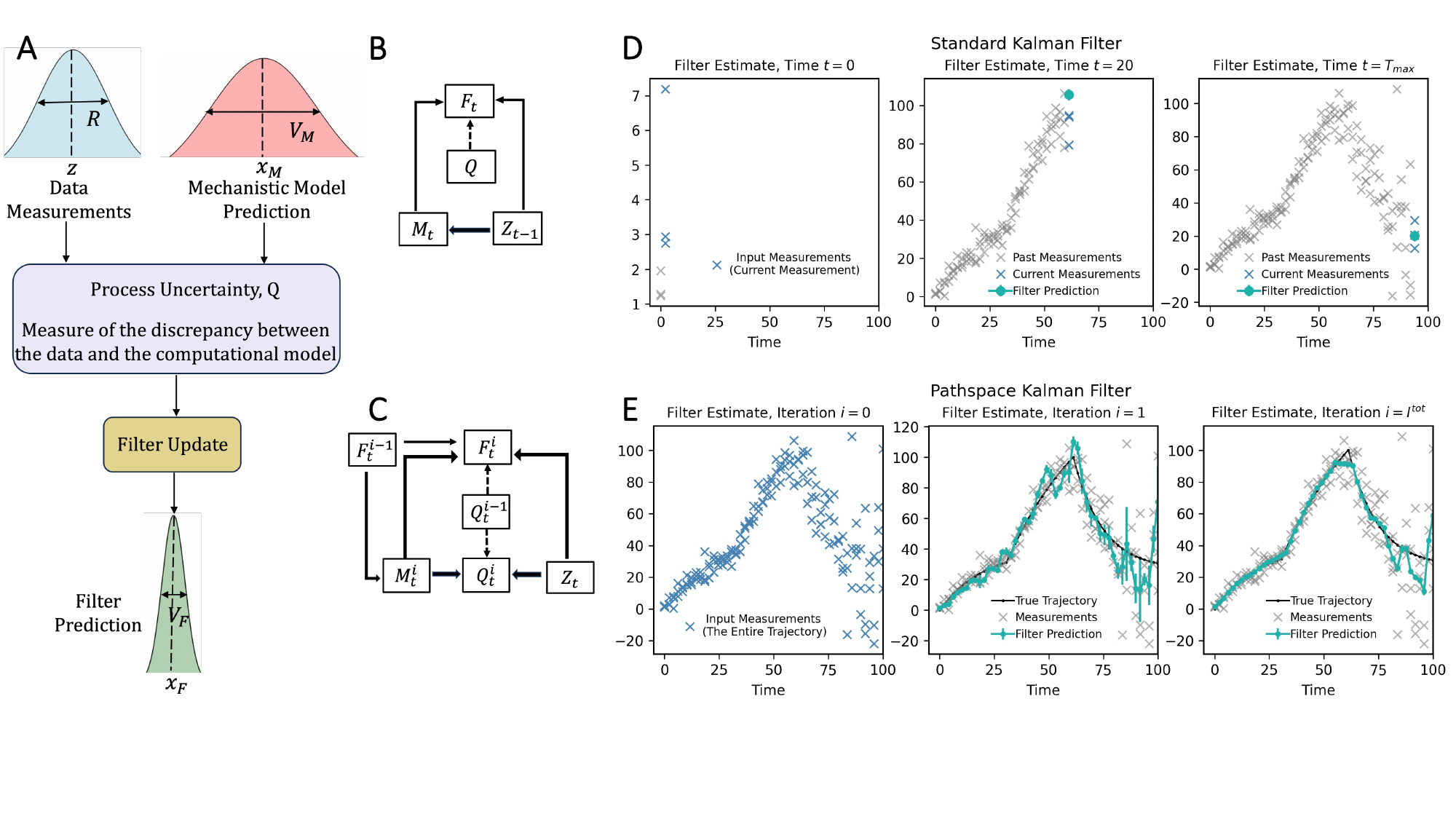}
    \caption{A. A schematic of the Kalman filtering framework. B. Graphical model for the KF. C. Graphical model of the proposed PKF. D. Illustration of a KF which makes predictions forward in time $t$ as new data is acquired. E. Illustration of a PKF which ingests the entire temporal trajectory, over all $t$, to produce an entire trajectory as output. At each iteration $i$, the PKF output trajectory is fed back into the PKF as the input for the next iteration $i+1$ of filtering.}
    \label{fig:graphicalabstract}
\end{figure*}

\noindent Estimating sources of uncertainty about a system from data is a quintessential problem with applications in many branches of science and engineering \cite{rao}, \cite{kf_hep}. The Kalman Filter (KF) is a provably optimal solution to this problem under certain assumptions \cite{kalman}. Crucially, KFs carefully account for different sources of noise, including data variance, model variance, and model confidence (called  process uncertainty) in order to yield optimal estimates. Figure \ref{fig:graphicalabstract}A illustrates the key components of a KF, and Figure \ref{fig:graphicalabstract}B and \ref{fig:graphicalabstract}D illustrate the dynamic state prediction of a standard KF. \\

\noindent The KF formalism has been extended in many ways including adaptive \cite{brown1985adaptive} and non-linear KFs \cite{extendedkf,unscentedkf} which allow diverse domain-specific internal models coupled with a general numerically tractable estimation procedure. This flexibility of KFs has allowed for its successful application in diverse fields such as economics \cite{harvey1990forecasting}, weather prediction \cite{galanis2006applications}, target tracking \cite{weng2006video}, data smoothing and assimilation \cite{NIPS1996_147702db}, flight \cite{5466132}, parameter estimation in electro-chemical responses \cite{BrownandRutan}, microscopy \cite{yang2006nuclei}, image processing \cite{nahata2023exploring}, music retrieval \cite{7042773}, and more.\\

\noindent More generally, KFs can be seen as simple probabilistic graphical models \cite{jordan2004introduction}. In this light, the KF models a system as a normal distribution with mean and variance derived as a convex combination of a measurement normal distribution and the internal model normal distribution, (Figure \ref{fig:graphicalabstract}A) allowing the relative contributions of noise from these sources to be quantified. Kalman Filters are traditionally used to iteratively predict the state of a system at a future time given a stream of incoming measurements (Figure \ref{fig:graphicalabstract}B, \ref{fig:graphicalabstract}E). \\

\noindent In many time-course experiments where KFs are used as an analysis tool, the entire measurement trajectory may be collected prior to analysis \cite{Jalles2009}, \cite{Lillacci2010,Huang2016}, \cite{Wu2002,Eden2008}. In such cases, Bayesian smoothing algorithms may improve state estimation by utilizing the entire trajectory. These algorithms frequently use KFs internally to iterate over the entire data trajectory and generate dynamic internal estimates of sources of uncertainty \cite{sarkka2023bayesian}. However, such approaches are sensitive to algorithm parameters which frequently must be predetermined by the user. \\

\noindent In the context of modeling biological phenomena as time-varying generative process, there should be no \textit{apriori} assumptions about the confidence in the underlying model being equal at all times times. In the KF formalism, this confidence is codified in the process uncertainty. In this work, we propose adaptively estimating the process uncertainty based on a loss function in order to dynamically quantify the reliability of the internal model. For example, if there is a change in the underlying data generation process which is not accounted for in the internal model, the process uncertainty should adapt to this discrepancy. These observations motivate the development of an algorithm which uses information from an entire trajectory (Figure (\ref{fig:graphicalabstract}C, \ref{fig:graphicalabstract}1D)) to improve its estimates. Our approach is to define a KF in pathspace which feeds trajectory estimates back into itself iteratively. We call our algorithm a Pathspace Kalman Filter (PKF), because, unlike a Bayesian smoother, our goal is not to smooth the data but rather to minimize state estimation variance and error across the entire trajectory. Finally, we note that an additional advantage of our approach is that it automatically allows for change point detection - meaning identifying when changes in a data generation process occur \cite{aminikhanghahi2017survey}. \\

\noindent \textbf{Contributions Summary:}
We make the following contributions to the existing standard Kalman Filter algorithm: \\

\noindent 1.  Pathspace Kalman Filter (PKF): In Section \ref{sec:PKF}, we describe the Pathspace Kalman Filter, which is a univariate non-linear adaptive KF that analyzes the entire time-series and dynamically quantifies process uncertainty while feeding its output trajectory (or path) back into itself, which we call iterating in pathspace. We prove the PKF allows for non-monotonic process uncertainty in time, yet that the PKF algorithm converges as it iterates in pathspace. \\

\noindent 2. We describe a scalable Bayesian methodology for simulation based computation of model expectation and variance in Section \ref{sec:bayesian_computation}. This allows the PKF to be run efficiently on large datasets such as transcriptomics data. \\

\noindent 3. Synthetic Data Experiments: In Section \ref{sec:experiments_main} we compare several KF algorithms and Bayesian smoothing algorithms with the PKF models on synthetic data and show that the PKF outperforms all other algorithms by one or more orders of magnitude in terms of mean-squared-error and requires less hyper parameter optimization. \\

\noindent 4. Real Data Experiments: In Section \ref{sec:gene_expression_main} we apply the PKF to a large time-course gene expression dataset with over 1.8 million measurements to demonstrate the utility and scalability of our method in the context of biological data. \\

\section{Background}
\label{sec:background}
\begin{table}[h]
\centering
\begin{tabular}{@{}lll@{}}
\toprule
\multicolumn{3}{c}{\small \textbf{\textsc{GENERAL}}} \\ \midrule
&$\E$        & Expectation value  \\
&$\V$        & Variance \\ \midrule
\multicolumn{3}{c}{\small \textbf{\textsc{ADAPTIVE NON-LINEAR KALMAN FILTER (KF)}}} \\ 
\multicolumn{3}{c}{\small \textit{(subscript $t$ indicates time)}} \\ \midrule
&$M_t$       & Model prediction \\
&$Z_t$       & Measurement values  \\
&$F_t$       & Filter output \\  
&$R$         & Data Variance (usually a constant) \\ 
&$Q$         & Process uncertainty (usually a constant) \\ 
&$w_t$       & Kalman Gain \\ 
&$m(Z_t)$    & Internal model of the KF \\ \midrule
\multicolumn{3}{c}{\small \textbf{\textsc{PATHSPACE KALMAN FILTER (PKF)}}}
\\
\multicolumn{3}{c}{\small \textit{(subscript $t$ indicates time, superscript $i$ indicates iteration)}} \\ \midrule
&$\M^i_t$    & Model prediction for PKF \\
&$\F^i_t$    & Pathspace Filter output \\ 
&$\V(Z_t)$   & Dynamic data variance \\ 
&$\Q^i_t$    & Dynamic Process uncertainty \\
&$\w^i_t$    & Data Weight \\ 
&$\wm^i_t$   & Model Weight \\ 
&$\wf^i_t$   & Weight of previous filter iteration\\
&$\m(\F^{i}, t)$    & Internal model of the PKF \\ 
\bottomrule
\end{tabular}
\caption{Frequently Used Notations}
\label{table:glossary} 
\end{table}

\subsection{Kalman Filters}
A 1-dimensional Kalman Filter (KF) is an optimal prediction-measurement-update algorithm in the sense that given a linear system with added Gaussian noise, the KF will produce the lowest variance estimate from the underlying data \cite{kalman}, \cite{kalmanfilternet}. Internally, a KF achieves this by linearly combining the predictions of an internal model $m$ with the observed data in order to fit a minimum variance normal distribution. In this work, we are interested in non-linear univariate KFs, so we do not require $m$ to be linear. Let $Z_{t} = Y_{t} + N_{t}$ be a sequence of data and $M_t \sim m(Z_{t-1})$ be a corresponding sequence of samples of model predictions with well defined means and variances. Here, $Y_{t}$ are the true values, $N_{t}$ are i.i.d. Gaussian noise, and $t$ indexing time-points, and $m$ is the internal model of the KF. Then the KF prediction is given by:
\begin{flalign}
    \label{eq:KF_update_1}
    \E(F_t) &= w_t \E(Z_{t-1})  + (1-w_t) \E(M_t) \\
    \label{eq:KF_update_2} 
    \V(F_t) &= w_t^2 \V(Z_{t-1}) + (1-w_t)^2 (\V(M_t) + Q) \\
    \label{eq:KF_update_3} 
    w_t &= \underset{w'}{\textrm{argmin}} [ w_t'^2 \V(Z_{t-1}) + (1-w_t')^2 (\V(M_t)+Q)]  
\end{flalign}
where, the terms are as described in Table \ref{table:glossary}. $\E(F_t)$ denotes the filter estimate of $Y_t$, which, in standard KFs is the minimum variance estimate of the process $Y_t$. 
The process uncertainty, $Q$, is traditionally used as a parameter to bias the relative weight between model and observations. Finally the weight $w_t$ is the Kalman gain which is monotonically decreasing in $t$ provided $Q$ and $\V(Z_t)$ are constant \cite{kalman}. In the case of a linear model $m$, this monotonicity gives rise to the convergence of the filter variance:
\begin{equation}
\V(F_t) = \V(F_{t-1})(1-w_t) \quad \textrm{with} \quad w_t \leq w_{t-1} \leq 1
\end{equation}
The Kalman gain represents new information gained with each subsequent measurement \cite{kalmanfilternet}. Note that in the original KF, the model $m$ is assumed to be a linear function. Extensions of this model such as the extended KF and unscented KF relax this constraint and allow for non-linear internal models \cite{extendedkf,unscentedkf}. In this text, we will generally use KF to mean a non-linear adaptive Kalman filter so that $m$ is not required to be linear and that the data variance $\V(Z)$ is dynamically updated.

\section{Pathspace Kalman Filter (PKF)}
\label{sec:PKF}

To motivate the PKF, we recognize that biological time-series analyses would benefit from an algorithm that is able to use an entire trajectory, looking forwards and backwards in time, to compute the process uncertainty dynamically. Such an algorithm should be able to highlight temporal windows where the data generative process deviates from our internal model. We define the PKF as an extension of a non-linear adaptive univariate KF where the entire filter iterates over its dynamical output (the path) repeatedly in a procedure similar to iterative Bayesian smoothing approaches such as the Iterated Posterior Linearization Smoothing (IPLS) algorithm \cite{IPLS}. Additionally, the PKF dynamically updates process uncertainty in order to re-weight the model confidence. \\

\noindent Our update equations are similar to a KF's with two conceptual changes. First, the filter's $i^{\textrm{th}}$ estimated trajectory distribution at time $t$, $\F^i_t$ (equation (\ref{eq:PKF_update_1})), is a convex combination of three Gaussian distributions instead of two: the data $(Z_t)$, the internal model prediction $(\M^i_t)$, and the previous filter update $(\F^{i-1}_t)$, which allows the filter to feed back on itself. Second, the PKF dynamically updates process uncertainty $(\Q^i_t)$ using a loss function $L$ (such as the $L_2$ norm) between the model output and the underlying data in order to minimize the loss (equation (\ref{eq:PKF_update_4})). Specifically the update equations are:
\begin{flalign}
    \label{eq:PKF_update_1}  
    \E(\F_t^i) &= \w^i_t \E(Z_t)  + \wm_t^i \E(\M^i_t) + \wf_t^i \E(\F^{i-1}_t) \\
     \label{eq:PKF_update_2}
    \V(\F_t^{i}) &= (\w_t^i)^2 \V(Z_t) + (\wm_t^i)^2 (\V(\M^i_t) + \Q_t^{i-1}) + (\wf_t^i)^2 \V(\F^{i-1}_t) \\
    \label{eq:PKF_update_3}
    \w^i_t, \wm^i_t &= \underset{\w'^i_t, \wm'^i_t}{\textrm{argmin}} \left[(\w'^i_t)^2 \V(Z_t) + (\wf'^i_t)^2 \V(\F^{i-1}_t) + (\wm'^i_t)^2 (\V(\M^i_t)+\Q_t^{i-1}) \right] \nonumber\\
    \textrm{\textit{s.t.}} \wf^i_t &= 1 - \w_t^i - \wm_t^i \\
    \label{eq:PKF_update_4}
    \Q_t^i &= \Q_t^{i-1} + (\w_t^i + \wm_t^i) (L(\M^i_t, Z_t) - \Q_t^{i-1}) .
\end{flalign}

\noindent There are three different weights $\w$, $\wm$ and $\wf$ for the data, model, and filter, respectively. The symbols have been made bold to denote that they are vectors in both time $t$ (subscript) and filter iteration $i$ (superscript). For example $\F^i = (\F^i_0,...,\F^i_T)$ with $T$ the number of time-points. Notice that the model samples $\M^i_t \sim \m(\F^{i-1}, t)$ are now a function of the same time $t$ and the entire trajectory $\F^{i-1}$ from the previous filter iteration instead of the single filter value $F_{t-1}$ at the previous time. Similarly, the process uncertainty $\Q_t^{i-1}$ is now a function of time and iteration instead of a constant. The prefactor $\w^i_t + \wm^i_t$ in equation (\ref{eq:PKF_update_4}) is analogous to a Kalman Gain for the process uncertainty and weights its update proportional to the additional uncertainty the model requires in order to match the data. 

\begin{algorithm}
\caption{Pathspace Kalman Filter}\label{alg:PKF}
\begin{algorithmic}
\STATE \textbf{Inputs:} data $Z$, model $\m$, iterations $I_{tot}$, loss function $L$, and total time points $T$
\STATE \textbf{Outputs:} $\E(\F^{I_{tot}})$, $\V(\F^{I_{tot}})$, and $\Q^{I_{tot}}$
\STATE $\E(\F_t^0) \gets \E(Z_t)$ \ALGCOMRIGHT{Initialization}
\STATE $\V(\F_t^0) \gets \V(Z_t)$
\STATE $\Q_t^0 \gets \V(Z_t)$ \\
\FOR{iteration $i=1$ \TO $I_{tot}$ }
    \FOR{timepoint $t=0$ \TO $T$ }
        \STATE $\M^i_t \gets \m(\F^{i-1}_t)$ 
        \ALGCOMRIGHT{Model Prediction}
        \STATE Compute $\w_t^i$, $\wm_t^i$, $\wf_t^i$ with Eq. (\ref{eq:PKF_update_3})
        \STATE Compute $\E(\F_t^i)$ with Eq. (\ref{eq:PKF_update_1})
        \ALGCOMRIGHT{Filter Update}
        \STATE Compute $\V(\F_t^i)$ with Eq. (\ref{eq:PKF_update_2})
        \STATE Compute $L(\M_i^t, Z_t)$ \ALGCOMRIGHT{Uncertainty Update}
        \STATE Compute $\Q_t^i$ via Eq. (\ref{eq:PKF_update_4})
\ENDFOR
\ENDFOR
\end{algorithmic}
\end{algorithm}

\noindent Feeding the filter output back into itself is important to ensure the convergence of the PKF as we prove in Appendix \ref{sec:Appendix_A2}. The PKF is illustrated as a graphical model in Figure (\ref{fig:graphicalabstract}C). Additionally, although it appears complex, the solution to equation (\ref{eq:PKF_update_3}) can be solved analytically as shown in equations (\ref{eq:PKF_weight1}-\ref{eq:PKF_weight3}) in appendix \ref{sec:Appendix_theorem2}. This entire construction can be viewed as an instantiation of the expectation maximization algorithm where in expectation step $i-1$ the filter output $\E(\F^{i-1})$ is computed. This value is then fed into the maximization step $i$ via the model $\M^i \sim \m(\F^{i-1})$ where the filter variance is minimized to compute $\w_t^i$, $\wm_t^i$ and $\wf_t^i$. The new Kalman gain can then be used to compute the filter output $\E(\F_{t}^{i})$ and so on. \\

\noindent In the KF formalism, the Kalman gain $\w_t$ can be interpreted as the amount of new information gained from the most recent measurement. This value decreases monotonically for a standard KF, meaning that the weighting of the uncertainty associated with the data can never increase in time \cite{kalmanfilternet}. Monotonicity of the Kalman gain in time is undesirable because it does not allow for dynamically varying (both increasing and decreasing) the relative contributions to uncertainties from the data and model over time. Next, we prove that for the PKF the Kalman Gain $\w^i_t$ is monotonically decreasing in $i$ but not $t$. \\

\noindent \textbf{Theorem 1}: The temporal sequence of Kalman Gains in the PKF, $\{\w^i_1, \w^i_2, ..., \w^i_t\}$ for any filter iteration $i$, as defined in equation (\ref{eq:PKF_update_3}) and updated according equations (\ref{eq:PKF_update_1} - \ref{eq:PKF_update_4}) may be non-monotonic in time $t$. \\

\noindent \textit{Proof Outline:} [Detailed proof in Appendix \ref{sec:Appendix_theorem1}]. For a given filter iteration, $i$, 
the Kalman Gain ($\w^i_t$) is given by:
\begin{equation}
    \w^i_t = \frac{AB} {AB + BC + CA}
\end{equation}
where, $A = \V(\F^{i-1}_t)$, $B = (\V(\M^i_t) + \Q^{i-1})$, and $C = \V(Z_t)$. As $\Q_t^{i-1}$ and $\V(\M_t^i)$ are not restricted to be monotonic in time, it follows that the sequence of weights, \{$\w^i_1$, $\w^i_2$,..., $\w^i_t$\} may be non-monotonic in time. \\

\noindent \textbf{Lemma of Theorem 1:} The sequence of process uncertainties $\{\Q_1^i, \Q_2^i, ..., \Q_t^i\}$ in time may be non-monotonic in time $t$. \\

\noindent \textit{Proof Outline:} [Detailed proof in Appendix \ref{sec:Appendix_theorem2}]. The non-monotonicity of $\Q^i_t$ originates in the non-monotonicity of the loss function $L$ referenced in Equation (\ref{eq:PKF_update_4}) when applied to dynamical data. The fact that the process uncertainty can vary over the course of a trajectory allows $\Q^i_t$ to be used as a dynamic metric to assess the relative sources of uncertainty from the model and data. In fact, as is apparent in equation (\ref{eq:PKF_update_4}), $\Q^i_t = L(\M_i^t, Z_t)$ is a fixed point of the PKF which quantifies a different error at each time-point. \\

\noindent  Next, we prove that the PKF algorithm converges to a fixed point in iteration space. \\

\noindent \textbf{Theorem 2}: In the limit of infinite iterations, the PKF converges to a fixed point of the process uncertainty $\Q^* = \lim_{i\to\infty} \Q^i$. \\

\noindent \textit{Proof Outline:} [Detailed proof in Appendix \ref{sec:Appendix_theorem2}].  First, we show that the filter variance is monotonically decreasing towards its fixed point $\lim_{i \to \infty} \V(\F_t^i)$. Then, we show the existence of a fixed point for $\lim_{i \to \infty} \Q_t^i$ and show that convergence of the filter variance guarantees that the process uncertainty also converges. Numerical results showing the convergence of the PKF can be found in Appendix \ref{appendix:numerical_convergence}. \\

\noindent An important feature of classical KFs is that in the linear case they minimize the means-squared error (MSE). In the next theorem, we show an analogous result for a class of PKFs that we call "accurate linear PKFs". Accurate linear PKFs have an internal model $\m$ that is a linear function which is identical to the underlying process dynamics. A formal definition is provided in Appendix (\ref{sec:Appendix_theorem3}).  \\

\noindent \textbf{Theorem 3}: An accurate linear PKF is optimal in the sense that it minimizes the MSE at each time-point $t$, and each iteration $i$. \\

\noindent \textit{Proof Outline:} [Detailed proof in Appendix \ref{sec:Appendix_theorem3}].  First, we show that filter output can be written as the expected value of the true process $(Y_t)$ with added Gaussian noise. Then, we prove that the accurate linear PKF update equations minimize the MSE. 

\section{Efficient Bayesian Model Computations}
\label{sec:bayesian_computation}
A challenge of iteratively running the PKF in pathspace is the necessity of many evaluations of the internal model $\m$. Notice that this model must generate a distribution of predictions at each time-point $\M_t^i$ with a well defined mean and variance. Additionally, we desire that our internal model be representative of the underlying process. Therefore, we parameterize the model predictions by solutions to an ordinary differential equation with varying underlying parameters $k$ and its solution over a time window $\Delta t = (t_0, t_f)$:
\begin{equation}
\label{eq:ode}
\frac{\textrm{d}x}{\textrm{d}t} =  g(x, k) \quad \implies x(k, \Delta t) = \int_{t_0}^{t_f} g(x; k) \textrm{d}t
\end{equation}
We define the subtrajectory $\zeta_{\Delta t}$ of values in the range $\Delta t$ as the input to the model $\m$. For $t \in \Delta t$, we define the PKF internal model as: $\m(\zeta_{\Delta t}, t)$ which is the distribution of the values of all paths $x(k, \Delta t)$ evaluated at time $t$. The probability $s(\zeta_{\Delta t}, k)$ of each path as parameterized by $k$ is given by:
\begin{equation}
\label{eq:posterior_dist}
s(\zeta_{\Delta t}, k) = \mathbb{P}(k \mid \zeta_{\Delta t}) \propto e^{-\mathcal{L}(x(k, \Delta t), \zeta_{\Delta t})} \pi(k) \\
\end{equation}
\noindent Here, $\pi(k)$ is a prior over the parameters and $\mathcal{L}$ is a loss function (such as the $L_2$ norm). In other words, $s(\zeta_{\Delta t}, k)$ is a distribution that assigns solutions of the ODE (\ref{eq:ode}) parameterized by $k$ a probability based on how well the solution fits all the data in the time window $\Delta t$. \\

\noindent There are three different weights $\w$, $\wm$ and $\wf$ for the data, model, and filter, respectively. The symbols have been made bold to denote that they are vectors in both time $t$ (subscript) and filter iteration $i$ (superscript). For example $\F^i = (\F^i_0,...,\F^i_T)$ with $T$ the number of time-points. Notice that the model samples $\M^i_t \sim \m(\F^{i-1}, t)$ are now a function of the same time $t$ and the entire trajectory $\F^{i-1}$ from the previous filter iteration instead of the single filter value $F_{t-1}$ at the previous time. Similarly, the process uncertainty $\Q_t^{i-1}$ is now a function of time and iteration instead of a constant. The prefactor $\w^i_t + \wm^i_t$ in equation (\ref{eq:PKF_update_4}) is analogous to a Kalman Gain for the process uncertainty and weights its update proportional to the additional uncertainty the model requires in order to match the data. \\

\begin{figure}
    \centering
    \includegraphics[width=\linewidth]{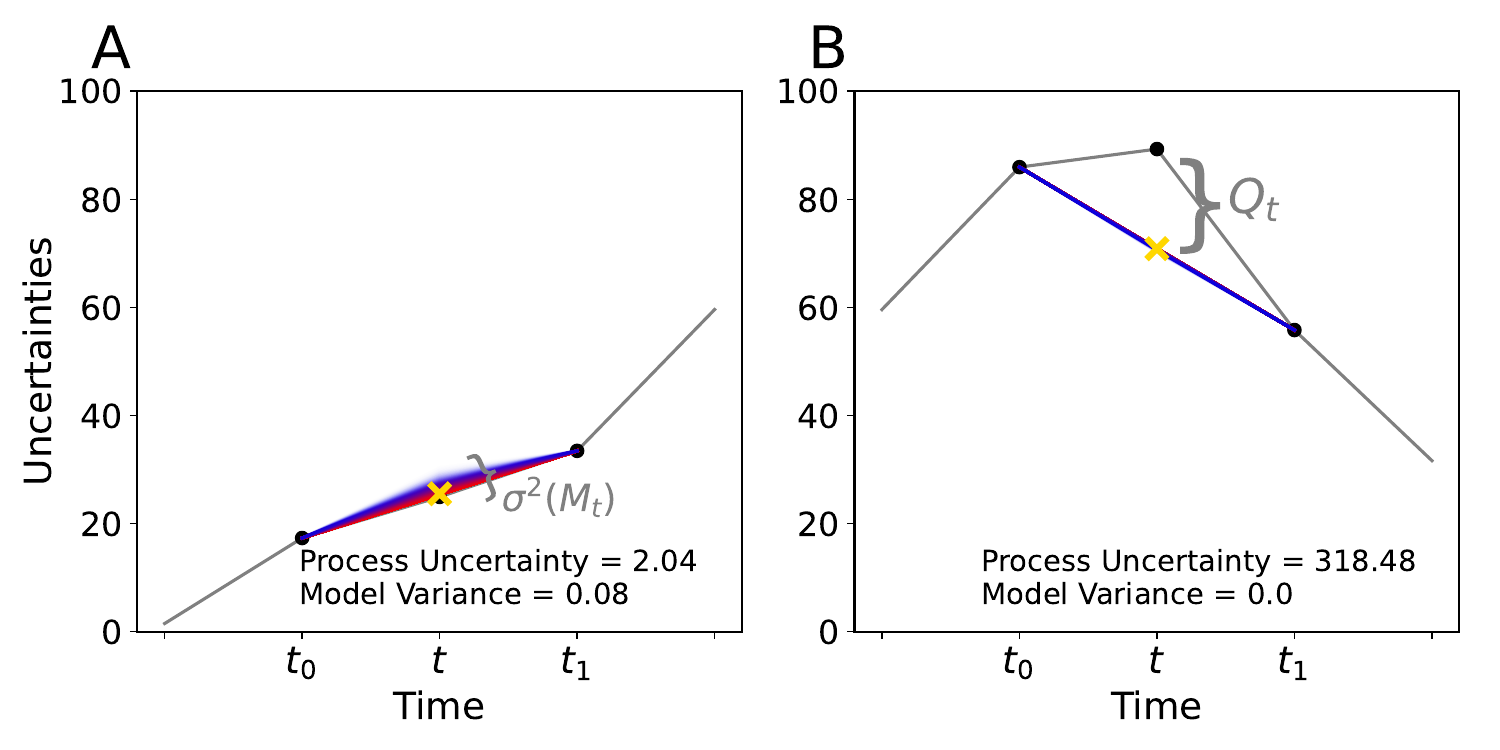}
    \caption{Models are fit across time windows of three measurements, where the first ($t_0)$ and third data-points ($t_1)$ are fixed, and the middle point (denoted by the yellow X mark at $t$) is allowed to vary. A. Illustrates the model fit and the process uncertainties when the model variance is relatively large, as represented by the spread of the colored lines, and the process uncertainty is small, as represented by the proximity of the estimated data-point (yellow X mark) to the measured data-point (black dot). B. Illustrates the fit where the process uncertainty is high, but the model variance is low. The implications of these relative measures are further elaborated in Table \ref{table:cases}.}
    \label{fig:bayesian_splines}
    \vspace{-0.5cm}
\end{figure}

\noindent In general, solving (\ref{eq:posterior_dist}) for high dimensions of $k$ can be numerically challenging but is possible using MCMC based approaches \cite{foreman2013emcee}. We simplify this problem by carefully choosing two parameter models $g(x, k_1, k_2)$ which can be analytically solved. Then, we choose $\Delta t$ to be a window of 3 consecutive time points. This allows us to derive an analytic function $h$ such that $k_2 = h(k_1, \zeta_{\Delta t})$ which derives one parameter from the other given the data. We then vary a single parameter $k_1$ (which unique determines $k_2$ via $h$) to analytically compute the distribution $s$. This procedure is illustrated graphically in Figure (\ref{fig:bayesian_splines}). Intuitively $s$ can be thought of as a local distribution of splines consistent with the ODE (\ref{eq:ode}). We describe this procedure more carefully in Appendix \ref{sec:Bayesian_spline_fit}. Different choices of $g$ are used for different kinds of data as described in the experiments section.

\subsection{Interpretation of the Bayesian Method and Comparison to Process Uncertainty}
\label{sec:interpretation_bayesian_methods}

In practice, the distribution of splines makes it easy to compute $\E(\M^i_t)$ and $\V(\M^i_t)$ by sampling $s(\zeta_{\Delta t}, k)$ via MCMC in high dimensional cases or by scanning over an array of values $k$ to compute the relevant statistics directly in low dimensional cases. Here, the variance represents how tightly the ODE model can be fit to the input data which is typically the previous filter output $\F^{i-1}$. By contrast, the process uncertainty $\Q_t^i$ is derived from the error between the model fit $\M_t^i$ and the underlying data $Z_t$. These two sources of uncertainty are illustrated in Figure (\ref{fig:bayesian_splines}). The total variance at each point of time will be a combination of the variance of the underlying model ($\V(\M^i_t)$), the process uncertainty ($\Q_t^i$), and the variance in the data to which the model is fit ($\V(\F_t^{i-1})$). Comparing the process uncertainty, the data variance, and the underlying model allows the PKF to detangle these different sources of uncertainties, which is infeasible using a traditional KF. We illustrate this analysis technique in the following section and Table \ref{table:cases}.

\section{Experiments}
\label{sec:experiments_main}

In order to validate the PKF algorithm, we compare to several varieties of KF and Bayesian smoothing algorithms using synthetic data because we needed access to the ground-truth state which is typically not present in real datasets. Then, to demonstrate the applicability of the PKF to real world problems, we used it to analyze a gene expression time-course dataset.

\subsection{Kalman Filters on Synthetic Data}
\label{sec:synthetic_data_main}

\begin{figure*}
    \centering
    \includegraphics[width=\linewidth]{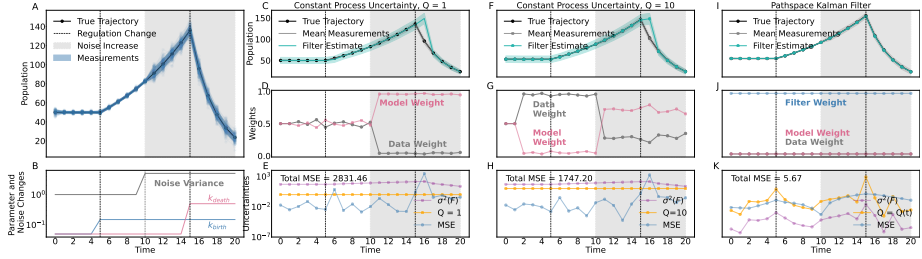}
    \caption{Comparison of the adaptive non-linear univariate KF and PKF on synthetic population dynamics data. A. A synthetic dataset of population dynamics generated from simulating a birth-death model and adding Gaussian noise. B. Parameters of the underlying model generating the simulated data. The growth and death rates change at the black vertical lines and the noise increases at the gray vertical line through the end of the time course, as indicated in all panels. C. Filter output for an adaptive non-linear KF with $Q=1$. Notice that the filter estimate lags changes in the data generating process. D. Model and data weights for the same KF. E. Process uncertainty, filter variance, and mean-squared error for the same KF. F-G. Filter output and parameters for an adaptive non-linear KF with $Q=10$. Notice that increasing $Q$ decreases model weight. H. Process uncertainty, filter variance, and mean-squared error for the same KF. I. Filter output for the PKF shows no lag. J. Weights for the PKF show that after many iterations, the filter converges to a low variance estimate because the filter weight $\wf > \w, \wm$. K. Mean-squared error, filter variance, and process uncertainty for the PFK. Notice that the mean-squared error is lowest for this model. Additionally, the process uncertainty spikes precisely when the data generating process changes.}
    \label{fig:synth_data}
\end{figure*}

\textbf{Data generation:} Synthetic population dynamics data were generated by simulating an ODE with time varying parameters and adding time-varying Gaussian noise to the output:
\begin{align}
\frac{\textrm{d}N}{\textrm{d}t} &= N(k_{birth}(t) - k_{death}(t)) 
\\
Z(t)  &\sim \mathcal{N}(\mu=N(t), \sigma = k_{noise}(t))
\end{align}
with regulation changes represented as time-varying rates:
\begin{align}
k_{birth}(t) &= 
    \begin{cases} 
      0.05 & t < 5 \\
      0.15 & t \geq 5
   \end{cases}
\quad
k_{death}(t) = 
    \begin{cases} 
      0.05 & t < 15 \\
      0.5 & t \geq 15
   \end{cases} \nonumber
\end{align}
and noise injection:
\begin{align}
k_{noise}(t) &= 
    \begin{cases} 
      1.0 & t < 10 \\
      5.0 & t \geq 50
   \end{cases} \nonumber
\end{align}
Here $N(t)$ is the ground truth population with time varying birth and death rates, $k_{birth(t)}$ and $k_{death}(t)$. The data $Z(t)$ are $100$ samples per time-point drawn from a normal distribution $\mathcal{N}$ with mean $N(t)$ and time varying standard deviation $k_{noise}(t)$. $Z(t)$ are fed into the the different Kalman Filter models. Mean-square-error is computed by comparing the filter outputs to the ground truth data $N(t)$. This synthetic data is plotted in Figure \ref{fig:synth_data}A and B.

\noindent The data generation was done with the objective to demonstrate the ability of the PKF to distinguish between the cases when the internal model is inaccurate with respect to the observed data (this occurs at times when the regulation changes), and when the data is noisy, but the internal model is reliable (which occurs when increasing noise is injected). Specifically, the PKF is able to distinguish the four cases described in Table \ref{table:cases}. \\

\noindent \textbf{Results for Synthetic Data:}
We compared the results PKF with several KF varieties and Bayesian smoothers on synthetic dataset described above. These results are summarized in Table \ref{table:method_comp} and plotted in Figure \ref{fig:method_comp}. The internal model used by all the Kalman filter variants also takes the form of a birth-death process, namely the ODE:
\begin{equation}
\frac{\textrm{d}x}{\textrm{d}t} = (k_{birth} - k_{death})x.
\end{equation}
The parameters $k_{birth}$ and $k_{death}$ are fit dynamically over windows of 3 time-points from the underlying data as described in Appendix \ref{sec:birth_death}. To compare the different KF outputs, we use mean-squared-error as a metric:
\begin{align}
     \textrm{MSE} = \frac{1}{T} \sum_{t=1}^T (\E(\F_t^i) - N(t))^2
\end{align}
where $T$ are the total number of time-points and either a KF or PKF can be used as the filter input into the expected value. \\

\noindent Columns 2 and 3 of Figure (\ref{fig:synth_data}) illustrate how a KF with different process uncertainties fit this dataset. Although increasing the process uncertainty decreases the model weight globally, the KFs both still increase the model weight when data becomes noisier. We remark that the Kalman gain (data weight) is not monotonically decreasing because the data variance is a dynamic parameter in this example. Similarly, the filter variance is also non-monotonic in this example due to the use of a non-linear internal model. \\

\noindent The results of the PKF are illustrated in Column 4 of Table (\ref{table:method_comp}). First, we note that the PKF has lower mean-squared-error than any of the other KFs at roughly 0.2\% of the KF MSE and roughly 10\% of the best alternative Bayesian smoothing algorithm. Second, we illustrate how the PKF is able to highlight temporal windows when the internal model's dynamics are insufficient to explain the data as illustrated in Figure (\ref{fig:synth_data}). Notice that changes in the data generation parameters as indicated by the black dashed lines result in instantaneous changes in the process uncertainties demonstrating how the PKF can be used to highlight time windows when the internal model cannot fit the data. However, the PKF is able to quickly adapt to these changes and recalibrate its internal model at future time-points. Indeed, even after the noise is increased by a factor of 10 (dashed gray line), the filter is still able to fit the data well. Together, these observations lead to the following heuristic of comparing the process uncertainty $\Q_t$ and data variance $\V({Z_t})$ as illustrated in Table \ref{table:cases}. \\

\noindent Case (A) in Table \ref{table:cases} represents the ideal situation where the internal model is highly predictive of reliable data. Case (B) in Table \ref{table:cases} represents a failure in the internal model and is an indication that additional temporal resolution or modeling efforts are needed in that time-window. Case (C) in Table \ref{table:cases} indicates that although the underlying data is noisy, the internal model is able to act as a reliable surrogate. Finally, in case (D) in Table \ref{table:cases} it is impossible to determine if the model inaccuracies are due to fitting noise or a lack of understanding about the underlying process. 

\begin{table}[h]
\begin{center}
\begin{tabular}{| M{10mm} | M{30mm} | M{30mm} | }
\hline
   & Low $\Q_t$ 
   & High $\Q_t$ \\ 
\hline
 \centering Low $\V(Z_t)$ 
 & A) Accurate Model \newline \& Reliable Data 
 & B) Inaccurate Model \newline \& Reliable Data \\  
\hline
 \centering High $\V(Z_t)$ 
 & C) Accurate Model \newline \& Noisy Data 
 & D) Inaccurate Model \newline \& Noisy Data \\
\hline
\end{tabular}
\caption{\label{table:cases} Interpreting different regimes of process uncertainty $Q_t$ and data variance $\V(Z_t)$.}
\vspace{-0.4cm}
\end{center}
\end{table}

\subsection{Application of the Pathspace Kalman Filter to Time-course Gene Expression Data}
\label{sec:gene_expression_main}
A prime motivation of this work is its application to biological time-courses, specifically to data obtained through high throughput acquisition methods such as bulk RNAseq and proteomics \cite{Wang2009,aebersold2003mass}. In these methods, high dimensional molecular snapshots of biological processes are collected at individual times via destructive experiments. For example, in RNAseq experiments the expression of  tens of thousands of genes may be measured simultaneously at each time-point but typically relatively few time-points are collected (usually at a resolution of several hours to days). The high dimensionality of this kind of data coupled with its temporal sparsity raises specific challenges. Firstly, biological systems are dynamic (not static) so one should expect them to change over time \cite{kitano2002systems}. A crucial question is whether the observed changes themselves are expected or unexpected; dynamic process uncertainty allows for the detection of discrepancies between observations and an underlying prior model. Secondly, the scalability of any analysis method to function on high dimensional data is crucial because a central goal of analyzing this kind of data is ascribing novel function to poorly understood genes, for example as targets for therapeutic interventions \cite{paananen2020omics}. In this section, we describe a model of gene expression suitable for detecting biologically important dynamics  and apply it to a time-course gene expression dataset. \\

\begin{figure}
    \centering
    \includegraphics[width=\linewidth]{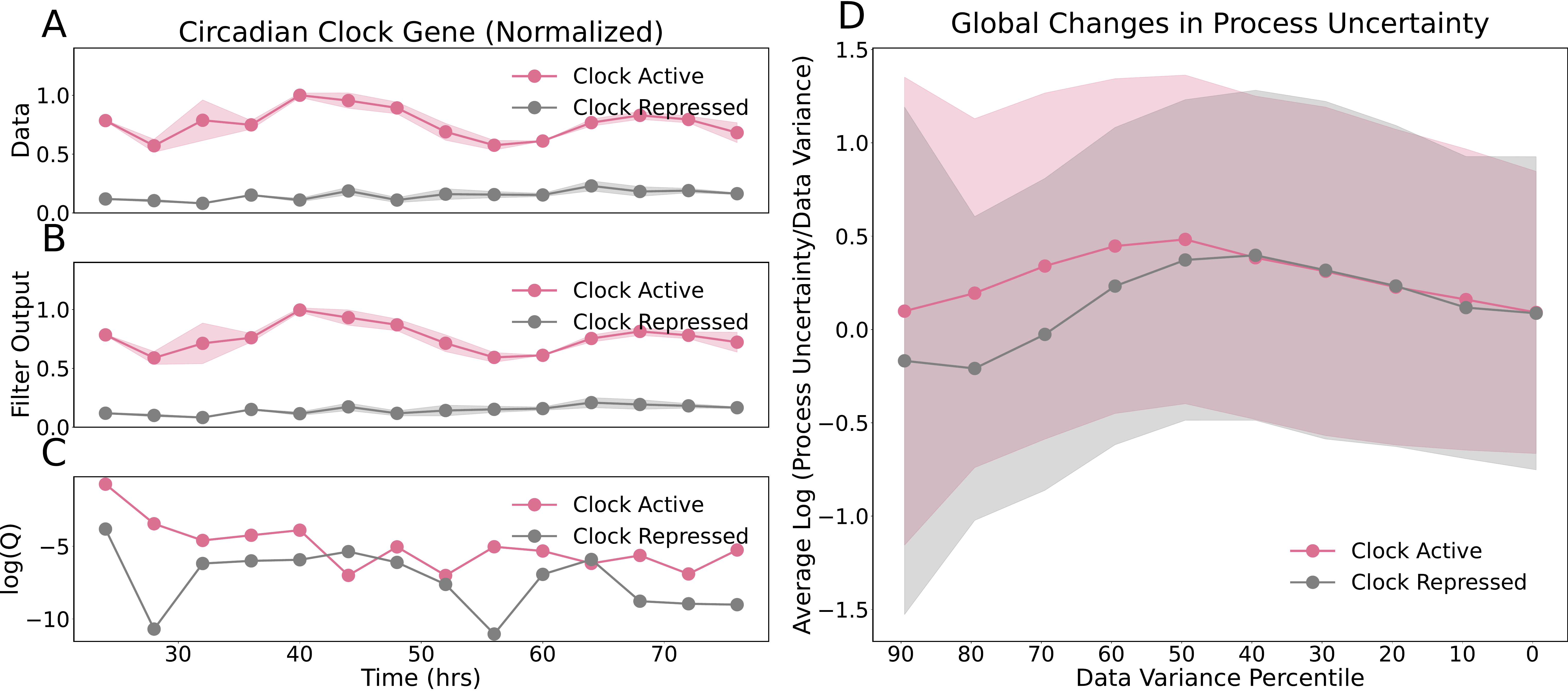}
    \caption{Application of the PKF to gene expression data in two conditions: Clock Active and Clock Repressed. A. Mean expression data of core circadian clock gene BMAL1 which oscillates when the clock is active and fails to oscillate when it is repressed. The shaded regions denote standard deviation. B. Filter output derived from the data in A. Shaded region denotes filter variance. C. Dynamic Process uncertainty associated with the filter output in B. D. Average log process uncertainty divided by data variance for $1.8$ million gene expression measurements separated by condition and plotted against their percentile variance. Turning off the circadian clock decreases process uncertainty for highly varying genes which is consistent with the circadian clock being a global regulator of gene expression.}
    \vspace{-0.5cm}
    \label{fig:circadian}
\end{figure}

\noindent \textbf{Dataset Description:}
In order to demonstrate the scalability of our approach on real world datasets, we used the PKF to analyze time-course gene expression measurements (RNAseq) of a circadian clock dataset in human cancer cell lines \cite{rebekah_clock} with contains of roughly 1.8 million gene expression measurements consisting of 32337 gene expression measurements across two conditions and 14 time-points with two replicates per time-point and condition. Briefly, the mammalian circadian clock is an cell-autonomous oscillatory genetic circuit which regulates diverse cellular functions by synchronizing them to daylight patterns \cite{takahashi}, \cite{sgolden}. For a brief description of the mammalian clock, refer to Appendix \ref{sec:gene_expression}. In this dataset, the circadian clock is disrupted by activating the gene MyC \cite{rebekah_clock} which causes the clock to be repressed. The effects of repressing the clock can be seen in Figure (\ref{fig:circadian}A) which shows the dynamics of the core circadian oscillator gene BMAL1. The failure of BMAL1 to oscillate indicates the repression of the circadian clock. Figure (\ref{fig:circadian}B) shows the filter output for BMAL1. \\

\noindent \textbf{Results for Real Gene Expression Data:}
In this exmaple, an internal model representing gene expression is used:
\begin{equation}
\frac{\textrm{d}x}{\textrm{d}t} = k_{exp} - k_{deg} x.
\end{equation}
Here, the expression rate ($k_{exp}$) and degradation rate ($k_{deg}$) of the transcript are assumed to be constant over short time windows--a hypothesis we call ``constant regulation" which we discuss in more detail in Appendix \ref{sec:gene_expression}. When the constant regulation model fails to accurately fit the data in a given window, it is indicative that the regulation of the gene is changing and is quantified by higher process uncertainty in the time window. This can be seen for BMAL1 in Figure (\ref{fig:circadian}C) where the process uncertainty is generally lower when the circadian clock is repressed. \\

\noindent Additionally, we ran the PKF on all genes in this dataset and compared the two conditions. The average log ratio of process uncertainty to data variance for these genes is plotted against variance percentile in Figure (\ref{fig:circadian}D). The circadian clock is known to regulate about 20\% of the genome \cite{takahashi}. Our analysis picks up this signal by showing increased process uncertainty divided by data variance for the top $50^{\textrm{th}}$ percentile of most varying genes when the circadian clock is active compared to when it is repressed indicating that regulation is less dynamic when the clock is deactivated. \\

\noindent \textbf{Complexity:} Finally, we note that the PKF with efficient Bayesian model computations is a highly scalable method and runs in time linear in the number of time-points, filter iterations, data dimensionality (e.g. number of genes), replicates and experimental conditions. We were able to run it on 1.8 million gene expression measurements on a single 8-core M1-PRO CPU in under an hour by parallelizing the algorithm on a per-gene basis.

\section{Related Work}
\begin{figure}
    \centering
    \includegraphics[width=\linewidth]{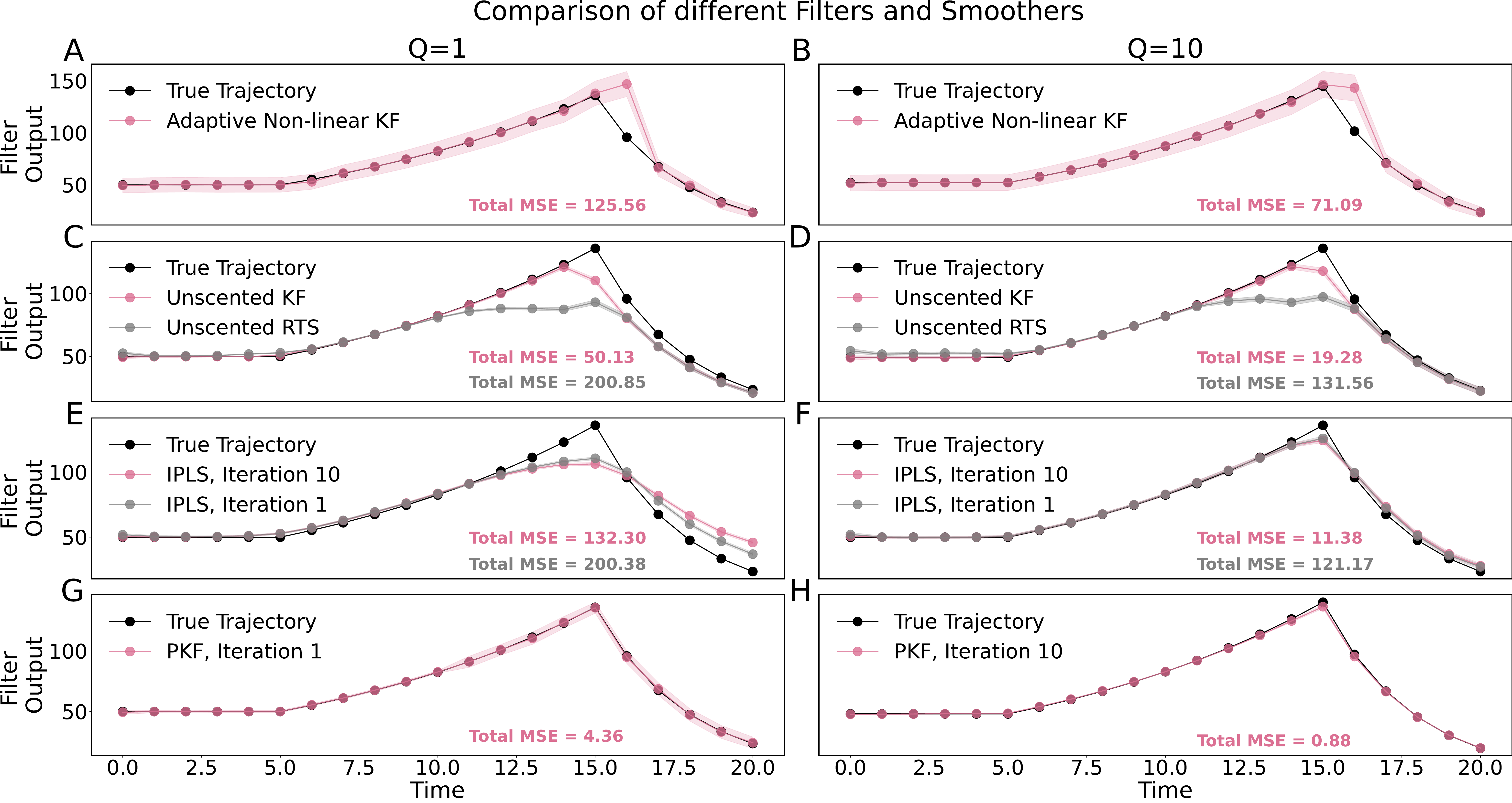}
    \caption{Outputs of the methods benchmarked in Table \ref{table:method_comp}.}
    \label{fig:method_comp}
    \vspace{-0.5cm}
\end{figure}

Table \ref{table:method_comp} and Figure \ref{fig:method_comp} compare several related methods to the PKF which outperforms all of them on MSE. We discuss these methods below. Additionally Table \ref{table:feature_table} in Appendix \ref{appendix:comparison} compares the various similarities and differences between the benchmarked algorithms. \\

\noindent \textbf{Extensions of the Kalman Filter:}
 Simple extensions include multivariate KFs \cite{multivariable_control}, non-linear or extended KFs \cite{extendedkf}, unscented KFs \cite{unscentedkf}, and ensembles of KFs \cite{ensemblekf}. Of particular relevance to our work are Adaptive KFs which dynamically adapts the filter's measurement model to the observed data by matching the covariances in the internal model predictions with the covariances of the measurements \cite{BrownandRutan}. Another inspiration for our work comes from analysis of the cyanobacteria circadian clock which is shown to be reminiscent of a Kalman filter non-monotonic Kalman gain \cite{Husain19}. Finally, KFs have been linked to various machine learning approaches\cite{Kim2022}. Several notable examples including using neural networks (NNs) to fit KF's process uncertainty and data variance parameters \cite{Greenberg2023}, using a NN as the KF's internal model \cite{Wu2012}, and replacing the Gaussian representation of distributions in the KF with normalizing flow NNs \cite{deBezenac2020}, and in hybrid learning models \cite{levine2022framework}. The PKF draws inspiration from many of these methods - particularly with regards to estimating process uncertainty dynamically - but is more closely related to Bayesian smoothing algorithms in its construction. \\
 \begin{table}
\begin{center}
\begin{tabular}{| c | c |c |}
\hline
  Algorithm & Parameters & MSE \\
  \hline
  \multirow{2}{*}{Adaptive Non-linear KF} & Q=1 & 125.56\\
  \cline{2-3}
  & Q=10 & 71.09\\
  \hline
  \multirow{2}{*}{Unscented KF} & Q=1 & 50.13\\
  \cline{2-3}
  & Q=10 & 19.28\\
  \hline
  \multirow{2}{*}{Unscented RTS} & Q=1 & 200.85\\
  \cline{2-3}
  & Q=10 & 131.56\\
  \hline
  \multirow{4}{*}{IPLS} & Q=1, Iterations = 1 & 200.38\\
  \cline{2-3}
   & Q=1, iterations=10  & 132.3\\
   \cline{2-3}
   & Q=10, Iterations=1 & 121.17 \\
   \cline{2-3}
   & Q=10, Iterations=10 &  11.38\\
  \hline
   \multirow{2}{*}{PKF} & Iterations=1 &  4.36\\
   \cline{2-3}
   & Iterations=10 &  0.88\\
\hline
\end{tabular}
\end{center}
\caption{Mean-square-error (MSE) of several methods on synthetic population dynamics data}
\label{table:method_comp}
\end{table}

\noindent \textbf{Comparison to Bayesian Smoothing:}
Bayesian smoothing algorithms are a class of smoothing algorithms with roots in Gaussian processes. These algorithms frequently use KFs ability to account for different noise sources in order to smooth data. The RTS smoother and unscented RTS smoother both use KFs to estimate states looking forwards and backwards in time across an entire trajectory \cite{RTS_smoother,UnscentedRTS}. The iterative posterior linearized smoother (IPLS) is an extention of the unscented RTS smoother which, similarly to a PKF, iterates the smoother output back into itself \cite{IPLS}. However, none of these methods dynamically update process uncertainty and, as we show in Table (\ref{table:method_comp}), this parameter can have dramatic effects on the mean-squared-error. The PKF solves this problem by automatically updating its process uncertainty as it iterates.
\noindent \textbf{Comparison to Parameter Estimation Algorithms:}
Parameter estimation is an immense topic, so here we focus on parameter estimation as it relates to KFs. Adaptive KFs attempt to estimate their internal parameters from data \cite{brown1985adaptive}. Many extensions of KFs have also been developed to estimate model parameters as observable state variables \cite{sarkka2023bayesian}. The PKF builds on these ideas by adaptively updating both its process uncertainty (via update equation \ref{eq:PKF_update_4}) and the internal model parameters via the internal Bayesian model computation described in section \ref{sec:bayesian_computation}. Unlike other approaches, the PKF uses analytically tractable internal models to directly compute Bayesian parameter posteriors which results in higher accuracy and more scalability. To our knowledge, no other methods iteratively update these parameters the way the PKF does. \\

\noindent \textbf{Comparison to Change Point Detection Methods:}
Change point detection methods are a large subject, so here we focus on methods that make use of KFs. Threshold based change point detection algorithms have been used to determine when to re-compute internal KF parameters \cite{PODDAR2016410_changepoint1}. Similarly to the PKF, the error of between filter estimates and data has been tracked to identify changes in the underlying model \cite{changepoint2}. Finally, ensembles of KFs have been combined with a Naive Bayes' algorithm for change point detection \cite{Daumer1998OnlineCD_changepoint3}. The dynamic process uncertainty of the PKF accomplishes a similar goal to these approaches but without the use of any hyper parameters such as thresholds. \\

\noindent \textbf{Methods to Analyze Time-course Biological Data:}
When using gene expression data, differential expression analysis is commonly employed \cite{anders2010differential}. However, this method does not explicitly take dynamics into account. KFs have been used to model gene regulatory networks by detecting statistically significant changes in time-varying linear KFs \cite{Xiong2013}  and using KFs to construct a Bayesian network \cite{Pirgazi2018}. A related approach is to fit a mechanistic model to the dynamical datasets which can be done with non-linear KFs \cite{Qian2008} or other methods \cite{Linden2022}. However, such approaches typically do not scale to large mechanistic models and suffer from issues related to parameter identifiability \cite{Lederer2024}.
 
\section{Conclusion}
\label{sec:discussion}

In summary, our model provides a powerful extension of KFs optimized for time-course datasets with the ability to dynamically highlight temporal regions where the model and data disagree. This method has been designed for use on biological datasets where identifying unexpected dynamics offers an opportunity for new scientific discovery. Towards this end, we have developed highly optimized Bayesian models for population dynamics and gene expression dynamics. We have applied these methods to real and synthetic datasets to demonstrate the accuracy and scalability of our approach. \\

\noindent \textbf{Limitations:} First, we comment that the PKF only deals with the univariate systems which could be extended to a multivariate formalism. Similarly, the assumption of normality of all underlying distributions could potentially be relaxed in future extensions of this method.\\

\noindent Second, fast parameter inference algorithms are necessary for our method to work on large datasets. We have presented a technique to implement these in cases where the internal model has a simple analytic solution. However, scaling up to large models remains an important challenge. Theoretically, black box Bayesian parameter inference methods could be used, but these will not scale well for large data applications such as those required in biology. \\

\noindent Although the PKF is undoubtedly more computationally intensive than some other filtering techniques, it is also more powerful and accurate. However, it is still fast enough to run on fairly large datasets without specialized compute resources and is therefore appropriate for many kinds of big data analyses. That said, for real-time data processing the PKF may not be appropriate.

\section*{Acknowledgements}
We thank the El-Samad, Bianco, and Multiscale Modeling groups at Altos Labs for their feedback on this project. We thank Joel Dapello, Vishrawas Gopalakrishnan, and Carlos Lopez for helpful feedback when preparing this manuscript.

\section*{Broader Impact, Ethical Implications, and Societal Consequences}
This paper presents work whose goal is to advance the field of Machine Learning. There are many potential societal consequences of our work, none which we feel must be specifically highlighted here.

\medskip
\printbibliography

\newpage
\appendix
\onecolumn
\section{Appendix}
\subsection{Proofs}
\label{sec:Appendix_A2}

\subsubsection{Proof of Theorem 1}
\label{sec:Appendix_theorem1}

\textbf{Theorem 1:} The temporal sequence of Kalman Gains in the PKF, $\{\w^i_1, \w^i_2, ..., \w^i_t\}$ for any filter iteration $i$, as defined in equation (\ref{eq:PKF_update_3}) and updated according equations (\ref{eq:PKF_update_1} - \ref{eq:PKF_update_4}) may be non-monotonic in time.\\

\noindent \textbf{Lemma 1.1} The temporal sequence of Process Uncertainties in the PKF $\{\Q_1^i, \Q_2^i, ..., \Q_t^i\}$ is not necessarily monotonic in time $t$.\\

\noindent \textit{Proof of 1.1:} This follows from the definition of $\Q_t^{i+1} = \wf_t^i \Q_t^{i} + (\w_t^i + \wm_t^i) L(M_t^i, Z_t)$. As $L(M_t^i, Z_t)$ is not required to be monotonic in time, it follows that the process uncertainty may not be as well. \\

\noindent \textit{Proof of Theorem 1:}
For a given filter iteration, $i$, the Kalman gain ($\w_t$) is given by:
\begin{equation}
    \w^i_t = \frac{\V(\M_t^i)+\Q_t^{i-1}}{\V(\M_t^i)+\Q_t^{i-1} + \V(\Z_t^i)}.
\end{equation}
By Lemma 1.1, $\Q_t^{i-1}$ is not restricted to be monotonic in time $t$. Similarly, the internal model $M_t^i$ can, in principle, be any function and is not restricted to be monotonic in time either. It follows that the sequence of weights, \{$\w^i_1$, $\w^i_2$,..., $\w^i_t$\} may not be monotonic in time, completing the proof. 

\subsubsection{Proof of Theorem 2}
\label{sec:Appendix_theorem2}

\textbf{Theorem 2}: In the limit of infinite iterations, the PKF converges to a fixed point of the process uncertainty $\Q^* = \lim_{i\to\infty} \Q^i$.\\

\noindent  \textbf{Lemma 2.1:} The iteration sequence of the PKF filter variances, for any given time-point $t$, $\{\V(\F^{(1)}_t), \V(\F^{(2)}_t), ..., \V(\F^i_t)\}$, as defined in equation (\ref{eq:PKF_update_2}) and updated according equations (\ref{eq:PKF_update_1} - \ref{eq:PKF_update_4}) is monotonically decreasing and approaches a fixed point $\V(\F_t^*) = \lim_{i\to\infty} \V(\F_t^i)= 0$. We begin by rewriting equations (\ref{eq:PKF_update_1}) and (\ref{eq:PKF_update_2}):
\begin{align}
\E(\F_t^i) &= \w_t^i \E(Z_t) + \wm_t^i \E(\M_t^i) + \wf_t^i \E(\F_t^{i-1}) \\
\V(\F_t^i) &= (\w_t^i)^2 \V(Z_t) + (\wm_t^i)^2 (\V(\M_t^i) + \Q_t^{i-1}) + (\wf_t^i)^2 \V(\F_t^{i-1}).
\end{align}

\noindent We minimize the variance of the PKF w.r.t. the weights $\w^i_t$ and $\wm^i_t$, with the constraint that $\w^i_t + \wm^i_t + \wf^i_t = 1$. Without loss of generality, we can choose any of the two weights for minimization. Here we choose to minimize the filter variance, one w.r.t. $\w^i_t$, and the second w.r.t. $\wm^i_t$, along with the constraint that $\wf^i_t = 1 - \w^i_t - \wm^i_t$.  \\

\noindent The first minimization w.r.t. $\w^i_t$ yields:
\begin{flalign}
    \frac{\textrm{d}(\V(\F^i_t))}{\textrm{d}\w^i_t} &=  0 \\ 
    \implies \w^i_t &= \frac{\V(\F^{i-1}_t)}{\V(\F^{i-1}_t) + \V(Z_t)} \left(1 - \wm^i_t \right)
    \label{eq:PKF_w1}
\end{flalign}
\noindent The second minimization w.r.t $\wm^i_t$ yields:
\begin{align}
    \frac{\textrm{d}(\V(\F^i_t))}{\textrm{d}\wm^i_t} &=  0 \\ 
    \implies \wm^i_t &= \frac{\V(\F^{i-1}_t)}{\V(\F^{i-1}_t) + \V(\M^i) + \Q^{i-1}_t} \left(1 - \w^i_t \right)
    \label{eq:PKF_w2}
\end{align}
\noindent Using equation (\ref{eq:PKF_w2}) in equation (\ref{eq:PKF_w1}), we have:

\begin{align}
    \w^i_t = \frac{\V(\F^{i-1}_t) (\V(\M^i_t) + \Q^{i-1}_t)}{\V(\F^{i-1}_t) (\V(\M^i_t) + \Q^{i-1}_t) + \V(\F^{i-1}_t) \V(Z_t) + \V(Z_t) (\V(\M^i_t) + \Q^{i-1}_t)} 
    \label{eq:PKF_weight1} \\
    \wm^i_t = \frac{\V(\F^{i-1}_t) \V(Z_t)}{\V(\F^{i-1}_t) (\V(\M^i_t) + \Q^{i-1}_t) + \V(\F^{i-1}_t) \V(Z_t) + \V(Z_t) (\V(\M^i_t) + \Q^{i-1}_t)} 
    \label{eq:PKF_weight2} \\
    \wf^i_t = \frac{(\V(\M^i_t) + \Q^{i-1}_t) \V(Z_t)}{\V(\F^{i-1}_t) (\V(\M^i_t) + \Q^{i-1}_t)+ \V(\F^{i-1}_t) \V(Z_t) + \V(Z_t) (\V(\M^i_t) + \Q^{i-1}_t)} 
    \label{eq:PKF_weight3}
\end{align}

\noindent To prove that the PKF converges in iteration space, we show that the filter variance is a recursive function of the previous iteration of the filter variance, multiplied by a factor less than 1. We do so by using the weights is equations (\ref{eq:PKF_weight1} - \ref{eq:PKF_weight3}) in equation (\ref{eq:PKF_update_2}). By defining  

\begin{equation*}
    A = \V(\F^{i-1}_t), \quad B = (\V(\M^i_t) + \Q^{i-1}_t), \quad C = \V(Z_t),
\end{equation*} 
we have:
\begin{flalign}
    & \V(\F^i) = \frac{A^2 B^2 C}{(AB + BC + AC)^2} + \frac{A^2 C^2 B}{(AB + BC + AC)^2} + \frac{B^2 C^2 A}{(AB + BC + AC)^2} \\
   & \V(\F^i_t) = \V(\F^{i-1}_t) \left( \frac{(\V(\M^i_t) + \Q^{i-1}_t) \V(Z)}{\V(\F^{i-1}_t) (\V(\M^i_t) + \Q^{i-1}_t) + \V(\F^{i-1}_t) \V(Z_t) + \V(Z_t) (\V(\M^i_t) + \Q^{i-1}_t)} \right) \\
\label{eq:VF_convergence}
    & \implies \V(\F^i_t) = \wf_t^i \V(\F^{i-1}_t) 
\end{flalign}
where $\wf_t^i \leq 1$ proving that the PKF converges monotonically in iteration space.\\

\noindent Next, we solve for the fixed point $\V(\F^*_t)$ using equation (\ref{eq:VF_convergence})
\begin{align}
\V(\F^*_t) = \wf_t^* \V(\F^*_t).
\end{align}
Here, $^*$ indicates the fixed point of $i$ after infinite iterations. This equation has two potential solutions: $\V(\F^*_t) = 0$ or $\wf_t^* = 1$. Analyzing the second case shows it is equivalent to the first:
\begin{align}
   & \wf_t^* = \frac{(\V(\M^*_t) + \Q^{*}_t) \V(Z_t)}{\V(\F^{*}_t) (\V(\M^*_t) + \Q^{*}_t)+ \V(\F^{*}_t) \V(Z_t) + \V(Z_t) (\V(\M^*_t) + \Q^{*}_t)} = 1 \\
   & \implies \V(\F^{*}_t) (\V(\M^*_t) + \Q^{*}_t)+ \V(\F^{*}_t) \V(Z_t) = 0 \\
   & \implies \V(\F^{*}_t) = 0
\end{align}
Generally speaking, the variances $\V(Z_t)$ and $\V(\M_t)$ are non-zero. Therefore this equation will have a single fixed point such that $\V(\F_t^*) = 0 \iff \wf^*_t = 1$. This also implies $\w_t^* = \wm_t^* = 0$ by the constraint from Equation (\ref{eq:PKF_update_3}). \\

\noindent \textit{Proof of Theorem 2:} We first show that $\Q_t^i$ has a fixed point $\Q_t^*$. Using equation (\ref{eq:PKF_update_4}):
\begin{align}
\Q_t^{*} = \wf_t^* \Q_t^{*} + (\w_t^* + \wm_t^*)L(\M_t^*, Z_t)
\end{align}
The previous lemma shows that $\wf^* = 1$ and $\w_t^* = \wm_t^* = 0$ which implies the existence of some $\Q_t^*$. \\

\noindent Next, we show that the PKF converges to this value of $\Q$. Lemma 2.1 shows $\V(\F_t^i)$ is monotonically decreasing towards 0 with $\wf_t^i$ approaching 1 with $\w_t^i$ and $\wm_t^i$ approaching 0. These three weights govern the updates for $\Q_t^i$ and their convergence will result in $\Q_t^i$ approaching its fixed point  $\Q_t^*$.  The convergence of the PKF is illustrated numerically in \ref{fig:convergence}.

\subsubsection{Proof of Theorem 3}
\label{sec:Appendix_theorem3}

First, we define an accurate linear PKF as a PKF which satisfies the following conditions:\\
1. The internal model of the PKF is linear scalar function of the form:
\begin{align}
    \M^i_t &= \m(X, t) = a_{t-1} X_{t-1} + b_{t-1} \quad \quad t > 0 \nonumber \\
    \M^i_0 &= \m(X, 0) = a_0^{-1} (X_{1} - b_0) 
\end{align}
where $a_{t-1}$ and $b_{t-1}$ are scalar.\\
2. The internal model of the PKF accurately captures the true environmental dynamics as a function of time, meaning:
\begin{equation}
    Y_{t} = a_{t-1} Y_{t-1} + b_{t-1}
\end{equation}
3. The filter is initialized to:
\begin{equation}
    \wf^0_t = 0 \quad\textrm{ and } \quad \M^0_t = Z_t.
\end{equation}
Furthermore, recall that the data $Z_t = Y_t + N_t$ where $N_t$ are i.i.d. samples from a normal distribution with mean 0. \\

\noindent \textbf{Theorem 3:} An accurate linear PKF is optimal in the sense that it minimizes the MSE at each time-point $t$, and each iteration $i$. \\

\noindent \textit{Proof of Theorem 3:} We prove this theorem in two different cases: one, with only measurement noise, and a second, with finitely many samples, and multiple sources of noise that propagate. Both these proofs rely on inductive reasoning, and minimization of the MSE w.r.t. the PKF weights. \\

\noindent \textbf{Case 1) Only Measurement Noise:} From equations (\ref{eq:PKF_update_1}-\ref{eq:PKF_update_4}), we have that:
\begin{equation}
    \E(\F^i_t) = \w^i_t \E(Z_t)+ \wm^i_t \E(\M^i_t) + \wf^i_t \E(\F^{i-1}_t)
\end{equation}
with $\w^i_t + \wm^i_t + \wf^i_t = 1$. First, we show that $\F^i_t = Y_t + \Gamma_t^i$, where $\Gamma$ is Gaussian white noise, and $Y_t$ is the true value of the underlying process by induction.\\

\noindent \textbf{Base Case} $i = 0$:
\begin{align}
    \F^0_t &= \w^0_t Z_t+ \wm^0_t \M^0_t \\
    \F^0_t &= \w^0_t (Y_t + N_t) + \wm^0_t Z_t \\
    \F^0_t &= (\w^0_t + \wm^0_t) Y_t + (\w^0_t + \wm^0_t + \wf^0_t) N_t \\
    \F^0_t &= Y_t + \Gamma_t^0
\end{align}
where $\Gamma_t^0 =  N_t$ is normally distributed with mean 0 by definition of $N_t$, and for this case, $\wf^0_t=0$, and $\w^0_t + \wm^0_t = 1$. \\

\noindent \textbf{Inductive Case:}
Assume $\F^i_t = Y_t + \Gamma_t^i$. \\
Show $\F^{i+1}_t = Y_t + \Gamma_t^{i+1}$. \\
We have:
\begin{align}
    \F^{i+1}_t &= \w^{i+1}_t Z_t +  \wm^{i+1}_t \M^{i+1}_t + \wf^{i+1}_t \F^{i}_t \\
    \F^{i+1}_t &= \w^{i+1}_t (Y_t + N_t) + \wm^{i+1}_t \m(F^{i}) + \wf^{i+1}_t \F^{i}_t \\
    \F^{i+1}_t &= \w^{i+1}_t (Y_t + N_t) + \wm^{i+1}_t (a_{t-1} F_{t-1}^i + b_{t-1}) + \wf^{i+1}_t \F^{i}_t  \\
    \F^{i+1}_t &= \w^{i+1}_t (Y_t + N_t) + \wm^{i+1}_t (a_{t-1} (Y_{t-1} + \Gamma_{t-1}^i) + b_{t-1}) + \wf^{i+1}_t (Y_t + \Gamma_t^i) \\
    \F^{i+1}_t &= Y_t (\w^{i+1}_t + \wm^{i+1}_t + \wf^{i+1}_t) + \wm^{i+1}_t a_{t-1} \Gamma_{t-1}^i + \w^{i+1}_t N_t + \wf^{i+1}_t \Gamma_t^i \\
    \F^{i+1}_t &= Y_t + \Gamma_t^{i+1}
\end{align}
Where $\Gamma_t^{i+1} = \wm^{i+1}_t a_{t-1} \Gamma_{t-1}^i + \w^{i+1}_t N_t + \wf^{i+1}_t \Gamma_t^i$ and is normally distributed with mean 0 because it is a sum of normally distributed variables with mean 0. \\

\noindent Next, we compute the MSE:
\begin{align}
    \E \left[ (\E(\F_t^i) - Y_t)^2 \right ] &= 
    \E \left[  \E(Y_t + \Gamma_t^i) - Y_t)^2  \right] 
    = \E \left[  (\E(Y_t) + \E(\Gamma_t^i) - Y_t)^2  \right]
    = \V(Y_t)
\end{align}
For a deterministic dynamical system, $\V(Y_t) = 0$ showing that the PKF optimally minimizes MSE produced by measurement noise. Similarly a system with non-zero variance, the $MSE$ of an estimator can never be better than the intrinsic noise in the system being measured. \\

\noindent \textbf{Case 2) Finite Measurement Samples:} For a measurement process $Z_t$ of an underlying process $Y_t$, with finite samples $n$, we note that the mean of the measurement process $\E(Z_t)$ will be centered around the expected true value $\E(Y_t)$, with a normally distributed error, with mean zero and variance $\eta^t = \V(Y_t)/\sqrt{n}$. This means:
\begin{equation}
    \E(Z_t) = \E(Y_t) + \eta_t
\end{equation}
Next we will prove a similar functional form for $\E(\M_t^i)$ and $\E(\F_t^i)$ by induction on $i$. \\

\noindent \textbf{Base case}: By definition, 
\begin{equation}
    \M_t^0 = Z_t \implies \E(\M_t^0) = \E(Z_t) = \E(Y_t) + \eta_t
\end{equation}
Then for the filter prediction:
\begin{align}
    \E(\F_t^0) &= \E(\w_t^0 Z_t + \wm_t^0 \M_t^0)\\
    \E(\F_t^0) &= \w_t^0 (\E(Y_t) + \eta_t) + \wm_t^0 (\E(Y_t) + \eta_t)\\
    \E(\F_t^0) &= \E(Y_t) + \eta_t,
\end{align}
completing the base case. \\

\noindent \textbf{Inductive Case:}
Assume:
\begin{equation}
    \E(\M_t^i) = \E(Y_t) + \xi_t^i 
    \textrm{  and  } 
    \E(\F_t^i) = \E(Y_t) + \gamma_t^i 
\end{equation}
where $\gamma_t^i$ and $\xi_t^i$ are normal distributions with mean 0. Then:
\begin{align}
    \E(\M_t^{i+1}) &= \m(F_t^i, t) \\
    &= a_{t-1}(\E(Y_{t-1}) + \gamma_{t-1}^i) + b_{t-1} \\
    &= \E(Y_t) + a_{t-1} \gamma_{t-1}^i = \E(Y_t) + \xi_t^i
\end{align}
where $\xi_t^i = a_{t-1} \gamma_{t-1}^i$ and is normally distributed with mean 0. Similarly,
\begin{align}
    \E(\F_t^{i+1}) &= \w_t^{i+1} \E(Z_t) + \wm_t^{i+1} \E(\M_t^{i+1}) + \wf_t^{i+1} \E(\F_t^i)
    \\
    \E(\F_t^{i+1}) &= \w_t^{i+1} (\E(Y_t) + \eta_t) + \wm_t^{i+1}(\E(Y_t) + \xi_t^{i+1}) + \wf_t^{i+1}(\E(Y_t) + \gamma_t^{i})
    \\
    \E(\F_t^{i+1}) &= \E(Y_t) + w_t^{i+1}\eta_t + \wm_t^{i+1} \xi_t^{i+1} + \wf_t^{i+1}\gamma_t^{i} = \E(Y_t) + \gamma_t^{i+1}
\end{align}
where $\gamma_t^{i+1} = \w_t^{i+1}\eta_t + \wm_t^{i+1} \xi_t^{i+1} + \wf_t^{i+1}\gamma_t^{i}$ and is normally distributed with mean 0 because it is a linear combination of such distributions. \\

\noindent Using these results, we will next show that the PKF update rules minimize the MSE for an accurate linear PKF.
\begin{align}
    &\E\left[ (\E(\F_t^i) - Y_t)^2 \right] = \E\left[ (\E(Y_t) + \gamma_t^{i} - Y_t)^2 \right] \\
    &= \E\left[
    \E(Y_t)^2 + (\gamma_t^{i})^2 + (Y_t)^2 + 2\E(Y_t)\gamma_t^{i} - 2\E(Y_t)Y_t - 2\gamma_t^{i+1}Y_t \right] 
    \\
    &= \E(Y_t)^2 + \E((\gamma_t^{i})^2) + \E(Y_t^2) + 2\E(Y_t)\cancelto{0}{\E(\gamma_t^{i})} - 2\E(Y_t)^2 - 2 \cancelto{0}{\E(\gamma_t^{i})}\E(Y_t) \\
    &= \E(Y_t^2) - \E(Y_t)^2 + \E((\gamma_t^{i})^2) \\
    &= \V(Y_t) + \E((\w_t^{i}\eta_t + \wm_t^{i} \xi_t^{i} + \wf_t^{i}\gamma_t^{i-1})^2) \\
    &= \V(Y_t) + (\w_t^i)^2 \E(\eta_t^2) + (\wm_t^i)^2 \E((\xi_t^i)^2) + (\wf^i_t)^2 \E(\gamma^{i-1}_t)^2 \nonumber \\
    & + 2 \cancelto{0}{\E(\eta_t)} \cancelto{0}{\E(\xi_t^i)} + 2 \cancelto{0}{\E(\eta_t)} \cancelto{0}{\E(\gamma_t^{i-1})} + 2 \cancelto{0}{\E(\xi_t^i)} \cancelto{0}{\E(\gamma_t^{i-1})}\\
    \E\left[ (\E(\F_t^i) - Y_t)^2 \right] &= \V(Y_t) + (\w_t^i)^2 \V(\eta_t) + (\wm_t^i)^2 \V(\xi_t^i) + (\wf^i_t)^2 \V(\gamma^{i-1}_t)
    \label{eq:minimize_mse}
\end{align}

\noindent The PKF is optimal only when these assumptions hold. In order for the algorithm to be optimal w.r.t. the MSE, we need to minimize Eqn (\ref{eq:minimize_mse}). First, we minimize w.r.t. $\w^i_t$, and then we minimize w.r.t. $\wm^i_t$. We then have:

\begin{align}
    & \frac{d}{d\w^i_t} \left(\V(Y_t) + (\w^i_t)^2 \V(\eta_t) + (\wm^i_t)^2 \V(\xi_t^i) + (\wf^{i}_t)^2 \V(\gamma_t^{i-1})\right) = 0  \\
     \label{eqn:theorem4_1}
    & \implies \w^i_t \V(\eta_t) = (1 - \w^i_t - \wm^i_t) \V(\gamma_t^{i-1}) \\
    & \frac{d}{d\wm^i_t} \left(\V(Y_t) + (\w^i_t)^2 \V(\eta_t) + (\wm^i_t)^2 \V(\xi_t^i) + (\wf^{i}_t)^2 \V(\gamma_t^{i-1}) \right) = 0 \\
    & \implies \wm^i_t \V(\xi_t^i) = (1 - 
    \w^i_t - \wm^i_t) \V(\gamma_t^{i-1}) 
    \label{eqn:theorem4_2}
\end{align}

\noindent Dividing Eqn (\ref{eqn:theorem4_1}) by Eqn (\ref{eqn:theorem4_2}), and after some algebra we have:
\begin{align}
    \w^i_t &= \frac{\V(\gamma_t^{i-1}) \V(\xi_t^i)}{\V(\gamma_t^{i-1}) \V(\xi_t^i) + \V(\gamma_t^{i-1}) \V(\eta_t) + \V(\eta_t) \V(\xi_t^i)}\\
    \wm^i_t &= \frac{\V(\gamma_t^{i-1}) \V(\eta_t)}{\V(\gamma_t^{i-1}) \V(\xi_t^i) + \V(\gamma_t^{i-1}) \V(\eta_t) + \V(\eta_t) \V(\xi_t^i)}\\
    \wf^{i}_t &= \frac{\V(\eta_t) \V(\xi_t^i)}{\V(\gamma_t^{i-1}) \V(\xi_t^i) + \V(\gamma_t^{i-1}) \V(\eta_t) + \V(\eta_t) \V(\xi_t^i)}
\end{align}

\noindent The weights that minimize the MSE are precisely the weights that we calculate in Appendix (\ref{sec:Appendix_theorem2}), with $\V(eta_t) = \V(Z_t)$, $\V(\xi^i_t) = \V(\M^i_t) + \Q^i_t$, and $\V(\gamma^{i-1}_t) = \V(\F^{i-1}_t)$, thereby proving that the accurate linear PKF is optimal w.r.t. the MSE.

\subsection{Numerical Convergence of the PKF}
\label{appendix:numerical_convergence}

\begin{figure}[h!]
    \centering
    \includegraphics[width=.5\linewidth]{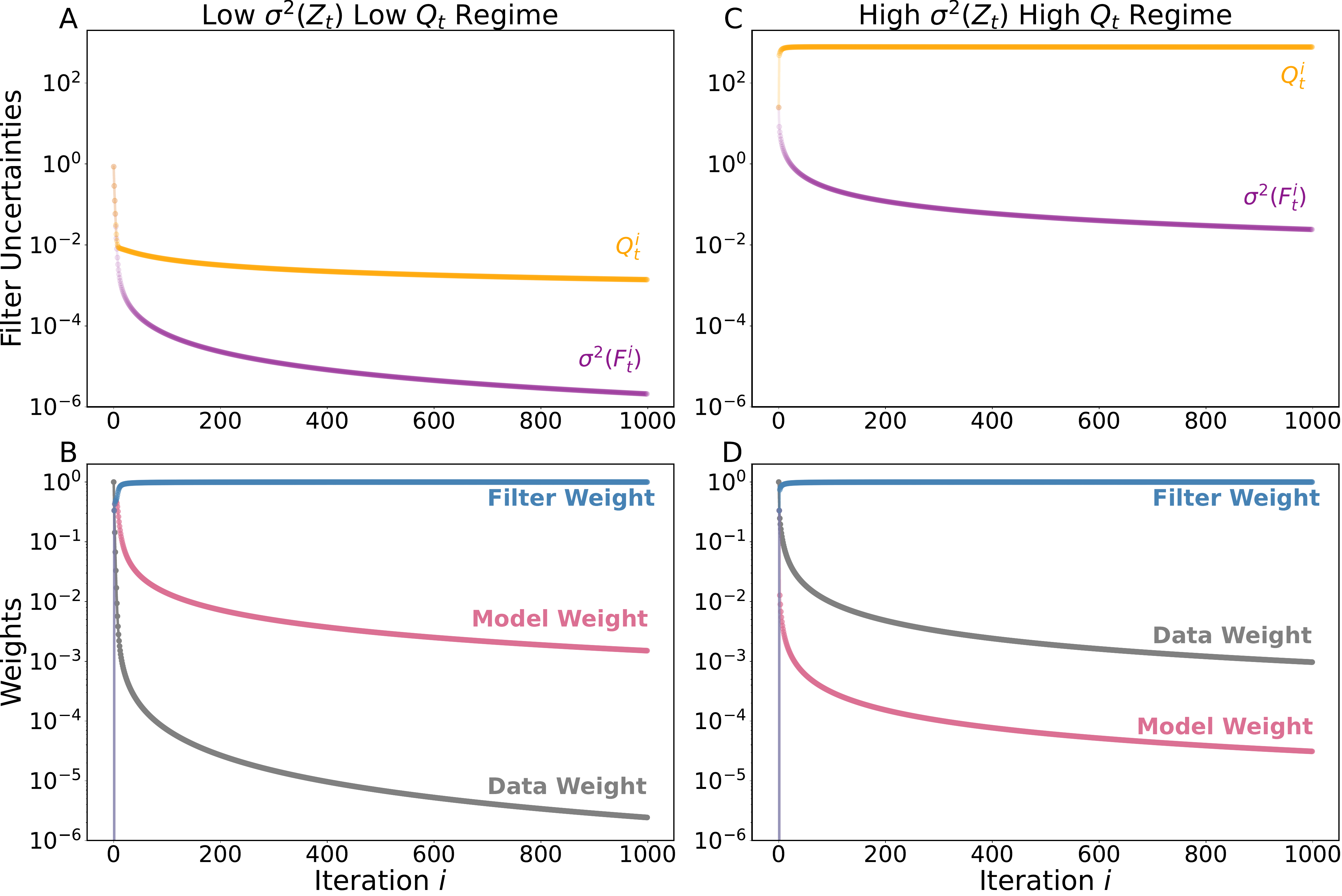}
    \caption{Convergence Plots for the PKF. For a time-point in the synthetic data where the Data Variance and Process Uncertainty are both low, Panel A shows Filter Variance and Process Uncertainty and Panel B shows the data, model, and filter weights as a function of iteration. C. For a time-point in the synthetic data where the Data Variance and Process Uncertainty are both high, Panel C shows Filter Variance and Process Uncertainty and Panel D shows the data, model, and filter weights as a function of iteration.}
    \label{fig:convergence}
\end{figure}

\noindent In Figure (\ref{fig:convergence}) in Appendix \ref{sec:Appendix_theorem2}, we show the numerical convergence of the algorithm on two different time-points and prove that the PKF converges. This is seen by the rapid convergence of $\Q_t^i$. The filter variance also converges towards 0, albeit very slowly. This behaviour may seem unexpected, however it occurs due to the feedback of the filter estimates into itself. In the limit of infinite iterations, the PKF has seen its own estimates increasingly more than it has seen the data or the model estimates which gives rise to decreasing filter variance. We emphasize that this behavior is not problematic as the most meaningful quantity in this framework is the process uncertainty and this converges to values representing the dynamic model uncertainty. 

\subsection{Monotonicity for alternative versions of the PKF} 
\label{sec:alternate_proof}

\begin{figure}[h]
    \centering
    \includegraphics[width=\linewidth,trim=3.5cm 3cm 3cm 3cm, clip]{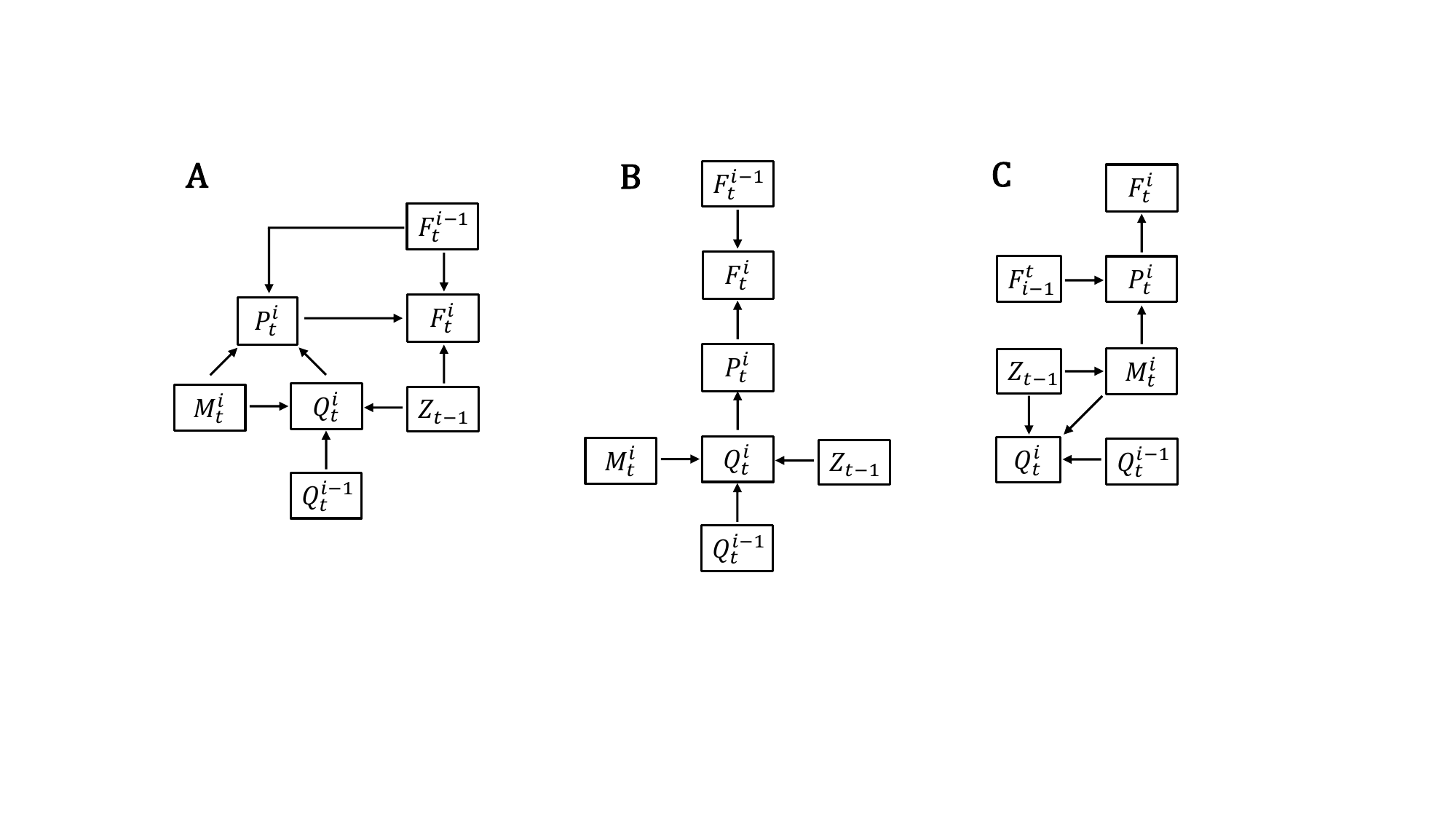}
    \vspace{-1.2cm}
    \caption{Graphical model illustrations of some alternate Pathspace Kalman Filter architectures.}
    \label{fig:alternate_architectures}
\end{figure}

\noindent The filters shown in Figure (\ref{fig:alternate_architectures}) are alternate version of the PKF used to obtain the results in the main manuscript. In this section, we will prove that the filter shown in (\ref{fig:alternate_architectures}A) converges and note that similar proofs hold for the other filter architectures in the figure. We highlight this point to demonstrate that there are a large class of pathspace Kalman filters with dynamic process uncertainty which converge. This convergence is always based upon feedback from the previous iteration of the filter into the filter, which can happen both directly and indirectly according to our numerical experiments (not shown). Understanding the nuances of different filter architectures is a potential avenue for future work. \\

\noindent In the version of the filter (\ref{fig:alternate_architectures}A), we have an additional layer of prediction, with the mean $\E(\pred)$ and variance $\V(\pred)$, which is anchored to the data and the internal model prediction. Then a convex combination of this prediction and the previous iteration filter output gives the final filter estimate. \\

\noindent Here we prove that this filter converges as well. We set up the filter as follows (suppressing the time index so we can use subscripts for variable names):
\begin{align}
    \E(\F^i) &= \w^i_P \E(\pred^i) + (1 - \w^i_P) \E(\F^{i-1}) \\
    \V(\F^i) &= (\w^i_P)^2 \V(\pred^i) + (1 - \w^i_P)^2 \V(\F^{i-1}) 
    \label{eq:alternate_filter_variance} \\
    \E(\pred^i) &= \w^i_1 \E(Z) + (1 - \w^i_1) (\E(\M^i) - \E(Z)) \\
    \V(\pred^i) &= (\w^i_1)^2 \V(Z) + (1 - \w^i_1)^2 (\V(\M^i) + \Q^{i-1}) 
    \label{eq:Prediction_variance}\\
    \Q^i_F &= \Q^i_M + \w^i_1 [(\E(M^i) - \E(Z))^2 - \Q^i_M] \\
    \Q^i_M &= h\left(\Q^{i-1}_F \right)
\end{align}

\noindent \textit{Proof:} To prove that this version of the PKF converges, we derive the weights $w^i_P$ and $w^i_1$, and show that in iteration space, the filter variance converges. The minimization of the prediction variance, as defined in equation (\ref{eq:Prediction_variance}) yields:
\begin{align}
    \frac{d (\V(\pred^i))}{d \w^i_1} &= 0 \\
    \w^i_1 &= \frac{\V(\M^i) + \Q^{i-1}}{\V(\M^i) + \Q^{i-1} + \V(Z)}
\end{align}

\noindent The minimization of the PKF variance yields:
\begin{align}
    \frac{d (\V(\F^i))}{d \w^i_P} &= 0 \\
    \w^i_P &= \frac{\V(\F^{i-1})}{\V(\F^{i-1}) + \V(\pred^i)}
    \label{eq:prediction_weight}
\end{align}

\noindent Using equation (\ref{eq:prediction_weight}) in equation (\ref{eq:alternate_filter_variance}), we have:
\begin{align}
    \V(\F^i) &= \frac{(\V(\F^{i-1}))^2 \V(\pred^i)}{(\V(\F^{i-1}) + \V(\pred^i))^2} + \frac{(\V(\pred^i))^2 \V(\F^{i-1})}{(\V(\F^{i-1}) + \V(\pred^i))^2} \\
    & \implies \V(\F^i) = \V(\F^{i-1}) k
\end{align}
\noindent where $k = \frac{\V(\pred^i)}{\V(\F^{i-1}) \V(\pred^i)} \leq 1$, proving that this alternate version of the PKF converges. \\

\subsection{Bayesian Optimization Framework}
\label{sec:Bayesian_spline_fit}
In this section, we describe the Bayesian optimization framework in more detail. Recall that we define the model output $\M_t^i = \m(\F^{i-1}, t)$ as the outputs from an ODE model:
\begin{equation}
\frac{\textrm{d}x}{\textrm{d}t} = g(x, k)
\end{equation}
with solutions from $t_0$ to $t_f$ given by the flow:
\begin{equation}
x(k, \Delta t) = \int_{t_0}^{t_f} g(x; k) \textrm{d}t.
\end{equation}
Here $\Delta t = (t_0, t_f)$. By varying parameters $k$ we can construct a distribution of flows (trajectories of the ODE) denoted:
\begin{equation}
\label{eq:sup:spline_prob}
    s(\zeta, \Delta t, k) = \mathbb{P}(k \mid \zeta) \propto e^{\mathcal{L}(x(k, \Delta t), \zeta_{\Delta t})} \pi(k)
\end{equation}
where $\mathcal{L}$ is a loss function, $\zeta_{\Delta t}$ is a vector of data in the range $(t_0, t_f)$ and $\pi(k)$ is a prior. In general, it is common to sample these kinds of distributions using a Markov Chain Monte Carlo (MCMC) in order to compute the so-called parameter posterior distribution \cite{Luengo2020}. In principle, a sufficiently expressive function $g(x; k)$ could be used with fixed parameters $k$ in order to represent the entire trajectory from the first to the last timepoint. However, as the dimensionality of $k$ increases, these distributions become increasingly challenging to sample efficiently. Instead our approach is to fit simple functions $g(x; k)$ with only two parameters $k = (k^{(1)}, k^{(2)})$ which can be analytically solved  exactly to fit data between two time points $t_0$ and $t_1$ with corresponding mean data $\zeta_0$ and $\zeta_1$. We call this fit an ODE spline and define it as:
\begin{align}
    & \overline{x}(k^{(1)}, h(k^{(1)}), t_0, t_1) = \int_{t_0}^{t_1} g(x;k)\textrm{d}t \\
    & \quad \textrm{s.t.} \quad \overline{x}(k^{(1)}, h(k^{(1)}), t_0, t_0) = \zeta_0 \\& \quad \textrm{and} \quad \overline{x}(k^{(1)}, h(k^{(1)}), t_0, t_1) = \zeta_1.
\end{align}
Here $k^{(2)} = h(k^{(1)})$ is analytically and uniquely determined by $k^{(1)}$ via the function $h$ derived from an analytic solution to $x$ so that the flow $\overline{x}$ precisely passes through the points $(t_0, \zeta_0)$ and $(t_1, \zeta_1)$.
\begin{figure}
    \centering
    \includegraphics[width=.4\textwidth]{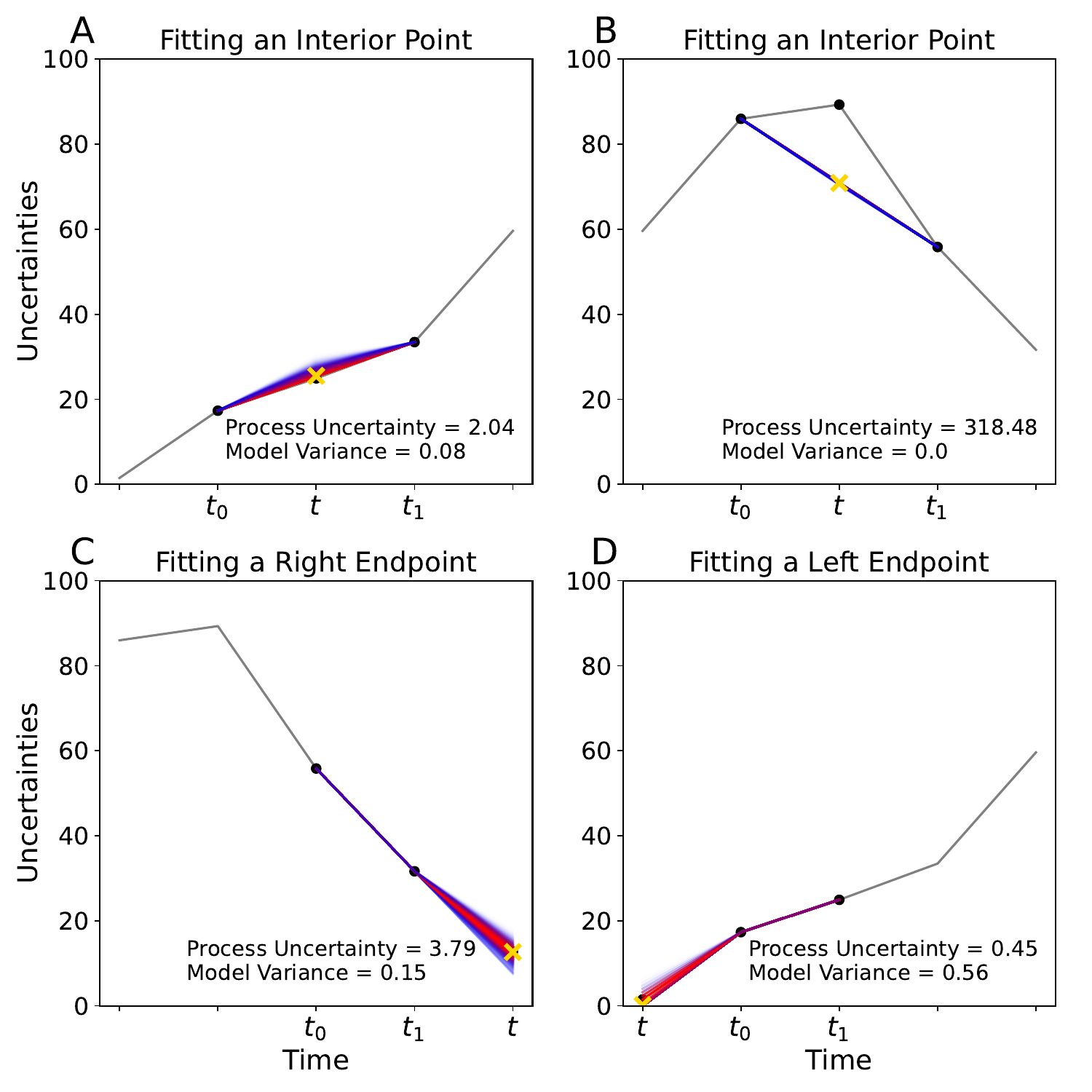}
    \caption{ODE Splines are fit by exactly fitting an ODE to two data points (black dots) and varying a parameter $k^{(1)}$ of the ODE to generate a distribution of fits for the third point (golden X). This process looks slightly different from interior points depicted in panels (A) and (B) versus right endpoint (C) and left endpoints (D).}
    \label{fig:spline_fig_appendix}
    \vspace{-0.5cm}
\end{figure}
We use the ODE splines $\overline{x}$ to compute distributions of flows by varying the parameter $k^{(1)}$ which in turn unique determines $k^{(2)}$. The weight (probability) of each spline is computed via Equation (\ref{eq:sup:spline_prob}) which will only take into account the loss at a third point $(t, \zeta_t)$ because the spline exactly fits the other points in the window. This procedure is illustrated graphically in Figure (\ref{fig:spline_fig_appendix}). Panels A and B show the case where $t_0 < t < t_1$ where an interior point is fit with the endpoints. Panel C shows the case of fitting a right endpoint, $t_0 < t_1 < t$ and panel D shows the case of fitting a left endpoint $t < t_0 < t_1$. We finally comment that this procedure is highly efficient due to the use of analytic formulas instead of MCMC sampling. Additionally, by computing the time-dependent parameter posterior probabilities, we can numerically integrate a distribution of different fits in order to compute model expected values and variances. The following sections illustrate this process on two different internal models.

\subsection{Analytics for the Numerical Experiments}
\label{sec:numerical_experiments}

In this section, we derive the expressions used for fast Bayesian computations of the custom internal models in the KFs we use for the main manuscript. 

\subsubsection{Growth-Death Model Analytics}
\label{sec:birth_death}

The simplest model (an example of $g$) for a birth-death process is given by the differential equation:
\begin{equation}
    \frac{d N(t)}{dt} = \left( k_{birth} - k_{death} \right) N_0.
\end{equation}
Note here we are using $N$ instead of $x$ to indicate that we are modeling a population. Solving this equation results in the flow ($\overline{G}$):
\begin{equation}
    N(t) = N(t_0) e^{\left(k_{growth} \right) \Delta t}
\end{equation}
\noindent where,  $\Delta t = t - t_0$, and $k_{growth} = k_{birth} - k_{death}$. 

\noindent For the spline-fit described in Appendix \ref{sec:Bayesian_spline_fit}, we need to fit the right end-point, the left end-point, and the center points. For each fit, we scan over an array of death rates $k_{death}$, and then compute the birth rates $k_{birth}$ via analytic functions $h$ which are slightly different for each fit-type (right, left, and center). \\

\noindent To fit the right end-point, we have:
\begin{align}
    \Delta t &= t_1 - t_0 \\
    k_{birth} &= h_{\textrm{right}}^{\textrm{birth-death}}(k_{death}) = k_{death} + \frac{1}{\Delta t} log \left(\frac{N(t_1)}{N(t_0)} \right) \\
    N(t_2) &= N(t_0) e^{k_{growth} (t_2 - t_0)}.
\end{align}

\noindent To fit the left end-point, we have:
\begin{align}
    \Delta t &= t_2 - t_1 \\
    k_{birth} &= h_{\textrm{left}}^{\textrm{birth-death}}(k_{death}) = k_{death} + \frac{1}{\Delta t} log \left(\frac{N(t_2)}{N(t_1)} \right) \\
    N(t_0) &= N(t_1) e^{k_{growth} (t_1 - t_0)}.
\end{align}

\noindent Finally, to fit the center-point, we have:
\begin{align}
    \Delta t &= t_2 - t_0 \\
    k_{birth} &= h_{\textrm{center}}^{\textrm{birth-death}}(k_{death}) = k_{death} + \frac{1}{\Delta t} log \left(\frac{N(t_2)}{N(t_0)} \right) \\
    N(t_1) &= N(t_0) e^{k_{growth} (t_1 - t_0)}.
\end{align}

\subsubsection{Gene Expression Model Analytics}
\label{sec:gene_expression}

In this section we describe the custom model used for the KF to analyze gene expression data. Here, we assume a different model $g$ which we call the constant regulation model, given by:
\begin{equation}
    \frac{d X(t)}{dt} = k_{exp} - k_{deg} X(t)
    \label{eq:gene_expression}
\end{equation}
\noindent where $k_{exp}$ is the rate of gene expression, $k_{deg}$ is the rate of degradation of the gene product. We call this a constant regulation model because it assumes that the expression and degradation rate of the gene are constant over a three-point time window. For short enough windows, we expect this to be true. However, if this model poorly fits the data, it suggests that the expression and/or degradation rates are being modulated by external factors in this time frame, ergo the regulation is not constant. The solution for equation (\ref{eq:gene_expression}) results in a flow $(\overline{G})$ given by:
\begin{equation}
    X(t) = \frac{k_{exp}}{k_{deg}} - \frac{X(t_0)}{k_{deg}} e^{k_{deg} (t - t_0)}
\end{equation}

\noindent For the spline-fit described in the above section, we need to fit the right end-point, the left end-point, and the center points. For each fit, we fix a range for the degradation rate $k_{deg}$, and then compute the expression rates $k_{exp}$ according to a function $h$ which is derived for to each fit (right, left, and center). \\

\noindent To fit the right end-point, we have:
\begin{align}
    \Delta t &= t_1 - t_0 \\
    k_{exp} &= h_{\textrm{right}}^{\textrm{const reg}}(k_{deg})  = k_{deg} \left(\frac{X(t_1) - X(t_0) e^{k_{deg} \Delta t}}{1 - e^{k_{deg} \Delta t}} \right)\\
    X(t_2) &= \frac{k_{exp}}{k_{deg}} - \left(\frac{k_{exp}}{k_{deg}} - X(t_0) \right) e^{k_{deg} (t_2 - t_0)}
\end{align}

\noindent To fit the left end-point, we have:
\begin{align}
    \Delta t &= t_2 - t_1 \\
    k_{exp} &= h_{\textrm{left}}^{\textrm{const reg}}(k_{deg}) = k_{deg} \left(\frac{X(t_1) - X(t_2)e^{k_{deg} \Delta t}}{1 - e^{k_{deg} \Delta t}} \right)\\
    X(t_0) &= \frac{k_{exp}}{k_{deg}} - \left(X(t_1) - \frac{k_{exp}}{k_{deg}}\right) e^{k_{deg} (t_1 - t_0)}
\end{align}

\noindent Finally, to fit the center point, we have:
\begin{align}
    \Delta t &= t_2 - t_0 \\
    k_{exp} &= h_{\textrm{center}}^{\textrm{const reg}}(k_{deg}) = k_{deg} \left(\frac{X(t_2) - X(t_0)e^{k_{deg} \Delta t}}{1 - e^{k_{deg} \Delta t}} \right)\\
    X(t_1) &= \frac{k_{exp}}{k_{deg}} - \left(\frac{k_{exp}}{k_{deg}} - X(t_0)\right) e^{k_{deg} (t_1 - t_0)}
\end{align}

\noindent Specifically, in Section \ref{sec:gene_expression_main}, we apply this framework to the RNAseq data as presented in \cite{rebekah_clock} As we want to focus on the regulation dynamics of the circadian clock, here we briefly describe how the mammalian circadian clock functions. 

\noindent The core circadian oscillator genes in the mammalian system are CLOCK, BMAL1, PER and CRY. Studies have shown that first BMAL1 and CLOCK form a hetero-dimer, CLOCK–BMAL1 complex, which then binds to regulatory elements containing E-boxes in a set of rhythmic genes that encode PER and CRY \cite{takahashi}. PER-CRY hetero-dimers act as the repressors of the BMAL1 and CLOCK genes, thereby achieveing the negative feedback required for sustained oscillations \cite{sgolden}. 

\section{Comparison of Various Smoothers and Filters}
\label{appendix:comparison}
In this Section, we compare the various state-of-the-art filters and smoothers, and discuss their implementations on the growth-death model described in Appendix \ref{sec:birth_death}. We compare the properties of different smoothing and filtering algorithms in Table \ref{table:method_comp}.

\subsection{Filter and Smoother Implementation Details}
\label{appendix:comparison_implementation_details}
We ran these algorithm on our synthetic growth dynamics data to compare to the PKF. Below we describe our process for implementing and running these comparison. All models made use of the same non-linear growth-death model:
\begin{equation}
    m(x, t_0, t_f; x_0) = \int_{t_0}^{t_f} x (k_{birth}(t) - k_{death}(t))dt \quad \textrm{with} \quad x(t_0) = x_0.
\end{equation}
Here, we inferred time-varying parameters $k_{birth}(t)$ and $k_{death}(t)$ using the method described in the Bayesian optimization framework section in appendix \ref{sec:Bayesian_spline_fit}. This ensures that all KF algorithms and smoothing algorithms had access to the same underlying non-linear model with the same parameters chosen to fit the data.
\begin{center}
\label{table:feature_table}
\begin{tabular}{@{} l cccccc @{}}
\toprule
  \multirow{5}{*}{Algorithm} & \multicolumn{6}{c}{Filter and Smoother Features} \\
  \cmidrule(lr){2-7}
  & \multicolumn{2}{c}{\thead{Input}} & \multicolumn{2}{c}{\thead{Process Uncertainty}} & \multirow{3}{*}{\thead{Parameter\\ Estimation}} & \multirow{3}{*}{\thead{Change-point\\ Detection}} \\
\cmidrule(lr){2-3}
\cmidrule(lr){4-5}
  & \thead{One\\ Time-point} &\thead{Full\\ Trajectory} & \thead{Constant\\ Vector} & \thead{Dynamically\\ Updated} & &  \\
\midrule
      \thead{Adaptive\\ KF} & \checkmark  & - & - & - & - & - \\ 
      \addlinespace
      \thead{Unscented\\ KF} & \checkmark & - & - & - & - & -  \\ 
      \addlinespace
      \thead{Unscented\\ RTS} & - & \checkmark & \checkmark & - & - & - \\ 
      \addlinespace
      \thead{IPLS} & - & \checkmark & \checkmark & - & - & -  \\ 
      \addlinespace
      \thead{PKF} & - & \checkmark & \checkmark & \checkmark & \checkmark & \checkmark  \\ 
\bottomrule
\end{tabular}
\end{center}

\subsubsection{Adaptive Non-linear Kalman Filter}
This model is closely related to the internal KF model used by the PKF. By adapting the PKF code to not feed back into itself (enforcing $\wf = 0)$, keeping process uncertainty $Q$ constant, and using the above right end point growth-death model instead of the center-point algorithm described, we arrived at an implementation of this baseline approach. Additionally, we call this algorithm \textit{adaptive} because the data variance $\V(Z_t)$ is recomputed at each timepoint based on the samples $Z_t$.

\subsubsection{Adaptive Unscented Kalman Filter}
The unscented transform is a commonly used technique to allow for flexible non-linear KFs \cite{unscentedkf}. We used a publicly available unscented KF implementation \cite{FilterPyPackage} with a custom internal model as described above. Additionally, just as in the Adaptive Non-linear KF, we the data variance $\V(Z_t)$ is recomputed at each timepoint based on the samples $Z_t$.

\subsubsection{Unscented Rauch Tung Striebel (RTS) Smoother}
The RTS smoothing algorithm is a classic Bayesian smoothing approach which, like the PKF, iterates over all the timepoints \cite{RTS_smoother}. The unscented RTS Smoother makes uses of the unscented transformation inside of an RTS smoother, functioning similarly to an unscented FK \cite{UnscentedRTS}. Unlike the PKF, the RTS does not feed its output back into itself. Instead, it uses a forwards and backwards prediction pass to propogate information along the trajectory. We used a publicly available implementation of the Unscented RTS smoother algorithm which internally used the same Adaptive Unscented Kalman Filter described above \cite{FilterPyPackage}.

\subsubsection{Iterative Posterior Linearization Smoother (IPLS)}
IPLS is the iterative Bayesian smoothing technique closely related to the PKF \cite{IPLS}. Both algorithms scan over a trajectory then feed their estimates back into themselves in an iterative fashion. However, the IPLS algorithm makes extensive use of sigma point linear regression to find optimal hyper parameters for an internal RTS smoother. Additionally, the IPLS is highly sensitive to the process uncertainty parameter $Q$ which must be supplied by the user and there are no clear criteria on how to choose this parameter.  The PKF solves this problem by dynamically fitting $Q$ to the data. We developed our own implementation the IPLS algorithm in python using the unscented Kalman Filter implementation described above internally.

\end{document}